\documentclass{article}
\usepackage{iclr2027_conference,times}
\iclrfinalcopy

\usepackage{amsmath,amsfonts,bm}

\def\eqref#1{equation~\ref{#1}}

\def\1{\bm{1}}

\DeclareMathAlphabet{\mathsfit}{\encodingdefault}{\sfdefault}{m}{sl}
\SetMathAlphabet{\mathsfit}{bold}{\encodingdefault}{\sfdefault}{bx}{n}

\usepackage[utf8]{inputenc}
\usepackage[T1]{fontenc}
\usepackage{amsmath,amssymb,bm,mathtools}
\usepackage{booktabs,graphicx,xcolor,colortbl,natbib}
\usepackage{wrapfig}
\usepackage{hyperref,url}
\renewcommand{\eqref}[1]{(\ref{#1})}
\newcommand{\worldwideweb}{\raisebox{-1.3pt}{\includegraphics[height=1.05em]{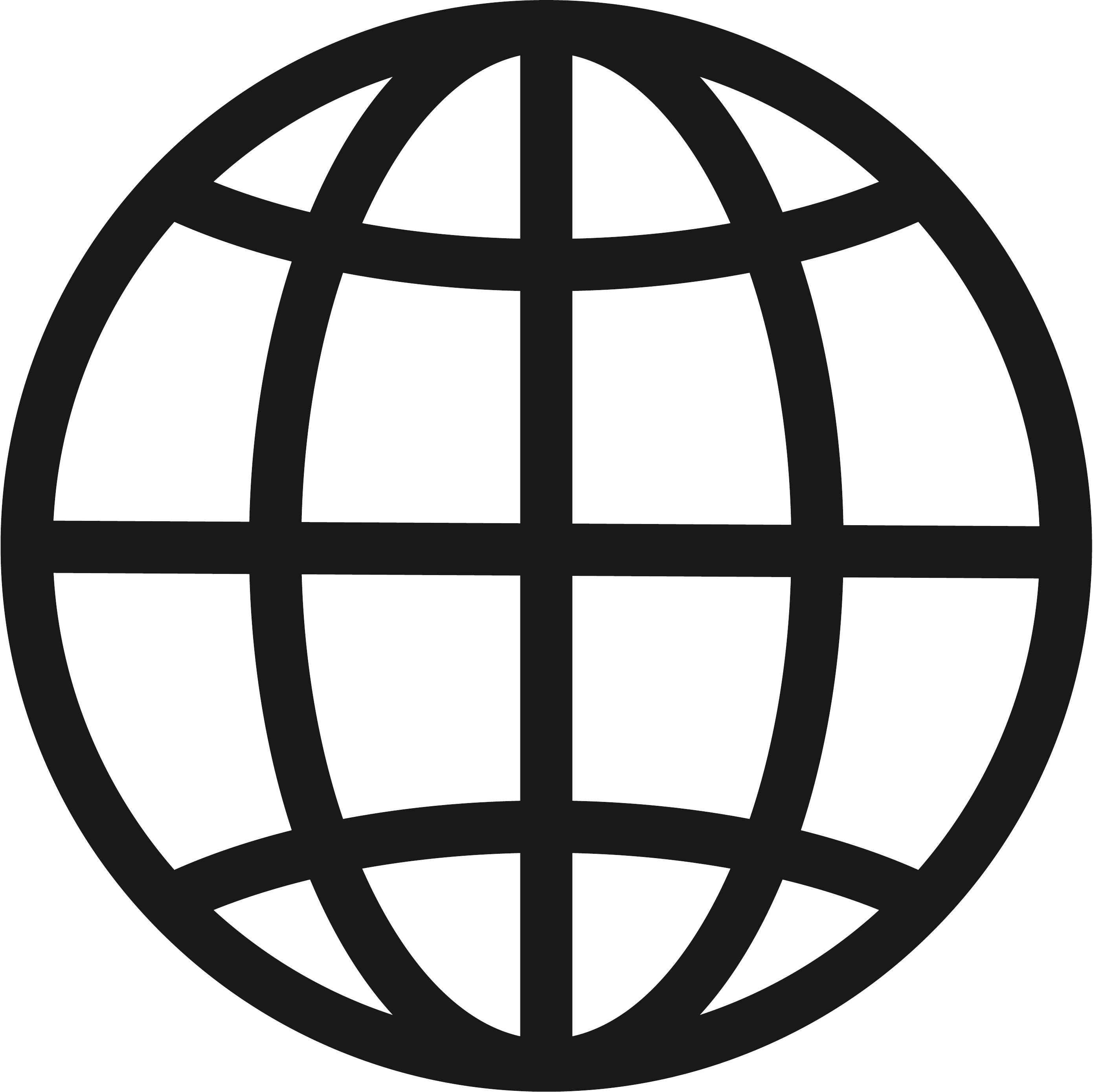}}}

\newcommand{\github}{\raisebox{-1.3pt}{\includegraphics[height=1.05em]{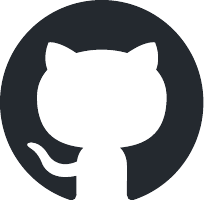}}}
\usepackage[table]{xcolor}   
\colorlet{lightred}{red!12}
\usepackage{tabularx}

\title{WIRE: Weight Space Implicit Representation Encoding of Frozen Video LLMs}
\author{
Ameen Ali$^{1}$, Tamim Zoabi$^{1}$, Lior Wolf$^{1}$ \& Chaim Baskin$^{2}$ \\
$^{1}$Tel Aviv University \\
$^{2}$INSIGHT Lab, Ben-Gurion University of the Negev, Israel \\
\texttt{\{ameenali,tamimzoabi\}@mail.tau.ac.il},~\texttt{wolf@cs.tau.ac.il} \\
\texttt{chaimbaskin@bgu.ac.il}
}
\hypersetup{
  hidelinks,
  pdftitle={WIRE: Weight Space Implicit Representation Encoding of Frozen Video LLMs},
  pdfauthor={Ameen Ali, Tamim Zoabi, Lior Wolf, Chaim Baskin}
}
\providecommand{\method}{\textsc{Wire}}
\newcommand{\benchcell}[2]{\rotatebox{90}{\parbox{2.0cm}{\raggedright\textbf{\scriptsize #1}\\{\scriptsize #2}}}}

\begin{document}
\maketitle
\fancyhead[L]{Preprint}

\begin{abstract}
Video language models represent a video as a long sequence of visual embeddings in
a decoder's input. Every visual embedding is processed by every decoder layer
during prefill, stored in the key value cache, and attended by every generated
token, so a long video raises prefill, memory, and generation cost at once.
We present \method{} (\emph{Weight Space Implicit Representation Encoding}), which
reduces this recurring cost by moving part of the video conditioned computation of a
frozen video language model from its input activations into a temporary low rank
update of its weights. \method{} keeps a short explicit visual context, so the
update acts as a correction rather than a replacement of the video. \method{} first
compresses the full resolution embeddings of
each frame into a small set of \emph{carrier tokens} using a question conditioned
bidirectional selective scan, and physically contracts the decoder input so every
decoder layer, cache entry, and decoding step operates on the shorter sequence.
Hypernetworks at selected decoder layers read the compressed forward pass together
with a readout of the full resolution features removed by contraction and generate
LoRA updates to the attention and feedforward projections. The updates are merged
after prefill, so autoregressive decoding runs the unmodified frozen backbone. On
Qwen3-VL-8B, \method{} with 16 carrier tokens per frame contracts the visual stream
by $3.5\times$, reduces decoder prefill time by $3.35\times$ at 256 frames and
$4.14\times$ at 1024 frames, shrinks the key value cache proportionally, and keeps
per token decoding latency approximately constant. Across nine
benchmark settings at 256 frames, it obtains the highest average accuracy
among efficient video language models we evaluate under a common 256 frame budget. A
paired update analysis toggles only the generated update while holding the
compressed visual context fixed. Applying the generated update raises whole benchmark
accuracy on Video-MME, MVBench, and LongVideoBench by up to $+4.4$ points, with the
largest gains on compression sensitive capabilities such as counting and action count,
establishing that the generated update does task relevant work beyond the carrier tokens.
\end{abstract}

    \begin{center}
{\normalfont\fontsize{10}{13}\selectfont
  \href{https://ameenali.github.io/wire/}{\worldwideweb~Project Page}
}
{\normalfont\fontsize{10}{13}\selectfont
  \href{https://github.com/AmeenAli/WIRE}{\github~Github Code}
}
\end{center}
\section{Introduction}
\label{sec:intro}
Video language models represent a sampled video as visual embeddings in
the input sequence of a decoder only language model~%
\citep{zhang2024llavavideo,bai2025qwen3vl,li2024llavaonevision,li2024videochatflash,
cheng2024videollama2,ye2024mplugowl3,liu2024kangaroo,zhang2025videollama3}:
a vision encoder maps each frame to a spatial grid of embeddings, interleaved
with timestamps and question text. This simple, effective interface has a structural
cost. Every visual embedding is processed by every
decoder layer during prefill, occupies a key value (KV) cache position, and is
attended by every output token, so a video that occupies thousands of
input positions inflates prefill compute, KV cache memory, and per token generation
cost all at once.

Existing efficient video language models cut this cost by pruning, merging,
resampling, or sparsifying the visual sequence, or by selecting frames~%
\citep{chen2024fastv,yang2025visionzip,islam2025bimba,shen2024longvu,tao2025dycoke,
ren2025vamba,song2025videonsa,qin2025videoxl2,jiang2025storm}.
These shorten the decoder input, but all retained visual information
still flows through the activation stream, is reprocessed by the full decoder, and
must contain enough to answer every question. At strong
compression, a compact visual sequence can lose information needed for counting,
action recognition, temporal ordering, and questions grounded in a particular
subtitle or event.

\method{} (\emph{Weight Space Implicit Representation Encoding}) explores a complementary
pathway, a temporary input conditioned weight update that carries a video conditioned
correction rather than a standalone representation of the video. Unlike activations, whose
cost grows with sequence length and output tokens, such an update is computed once and merged
before decoding.

Let $V$ denote a video, $q$ its question, $X$ the original multimodal decoder
input, $\mathcal{C}$ a visual context compressor, and $f_\theta$ a frozen video
language model. \method{} learns a compressor and a video question conditioned
update $\Delta\theta(V,q)$ such that
\begin{equation}
 f_{\theta+\Delta\theta(V,q)}\bigl(y\mid\mathcal{C}(X)\bigr)
 \approx f_\theta\bigl(y\mid X\bigr).
 \label{eq:objective}
\end{equation}
The contracted input $\mathcal{C}(X)$ retains a small, spatially organized set of visual
\emph{carrier tokens}, and the generated update changes how the frozen decoder processes
them and the question.

\begin{figure*}[t]
\centering
\includegraphics[width=\textwidth]{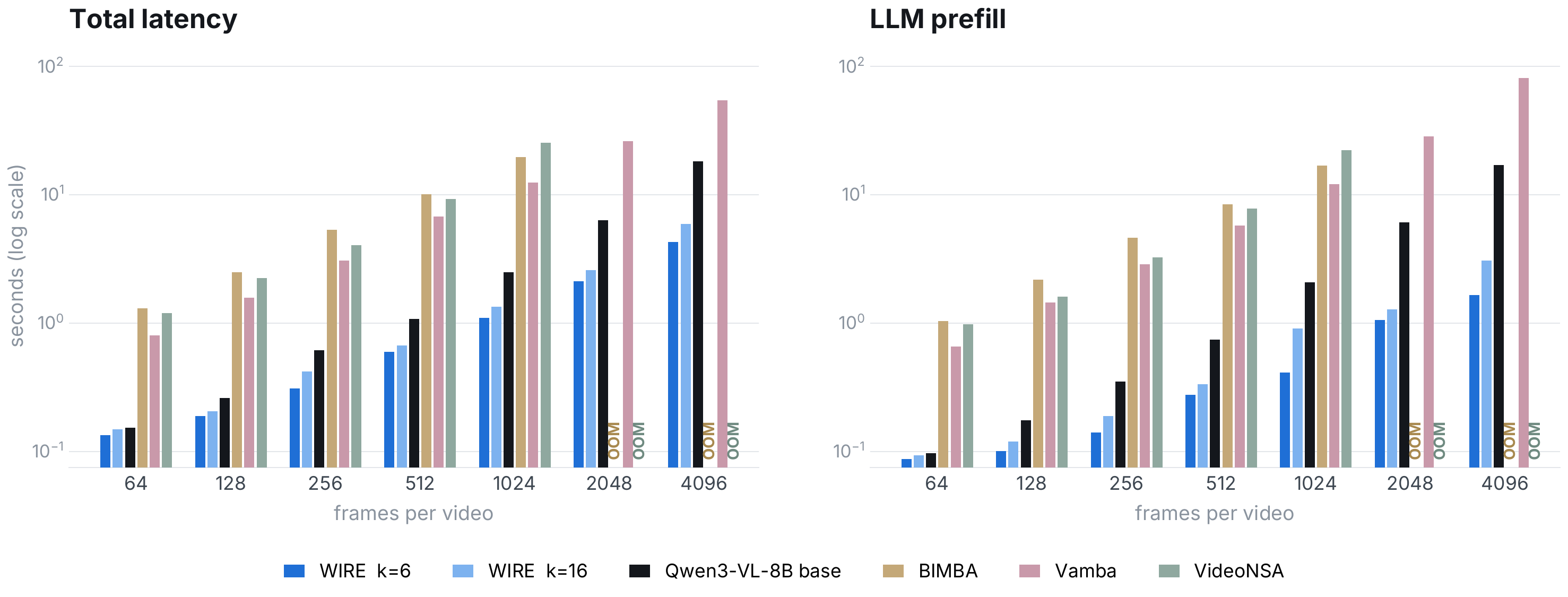}
\caption{\textbf{Latency versus video length.} Times are measured on one B200 GPU
at batch size one in bfloat16, excluding video file decoding. \emph{Left:} total
model time to first token. \emph{Right:} decoder prefill only. \method{} contracts
the sequence before the first decoder layer, so its advantage grows with frame
count and it continues to run beyond input lengths where several baselines exhaust
memory.}
\label{fig:latency}
\end{figure*}

Our contributions are threefold. First, we introduce a frozen video LLM
architecture that pairs physical visual token contraction with per example
weight space adaptation, so sequence length reduction reaches every decoder
layer, cache entry, and decoding step, separating \method{} both from methods that keep all
retained evidence in the activation stream. Second, we train the
compressor and hypernetworks by context distillation~\citep{hinton2015distilling,snell2022context}
from the same frozen backbone on the uncompressed input. Third, we provide an accuracy versus
latency frontier across nine video benchmark settings and a large paired update analysis
(Section~\ref{sec:results-weight}) that toggles only the generated update at fixed compressed
context, measuring whether it does task relevant work beyond the carrier tokens; whether that
work is video specific we leave as a hypothesis. Figure~\ref{fig:latency} shows that because
\method{} contracts the sequence before the first decoder layer, its prefill and time to first
token advantage widens with frame count.
\section{Related Work}
\label{sec:related}

\paragraph{Efficient video language models.}
Most efficient video language models shorten the decoder sequence by selecting, merging,
or pooling visual
embeddings~\citep{chen2024fastv,yang2025visionzip,huang2025prunevid,tao2025dycoke,
liu2026infomerge,bolya2023tome,zhang2025sparsevlm,xing2025pyramiddrop,shao2025tokencompression},
by resampling into a few carrier
tokens~\citep{jaegle2021perceiver,alayrac2022flamingo,li2023blip2,mu2023gisting,petrov2025gistpool},
or by adapting spatiotemporal token allocation under a
budget~\citep{ye2025vocollama,shen2024longvu,cai2025matryoshka,zhang2026xcomp,
zohar2025apollo,liu2024nvila,zhang2025llavamini,li2024llavaonevision}. A second family
keeps the visual sequence but cuts its attention or key value cache cost, sparsifying
visual attention or compressing the cache
mid decoder~\citep{song2025videonsa,tu2025vlcache,lin2025vtw,wang2025adaretake,
qin2025videoxl2,xiao2024streamingllm,zhang2023h2o,kim2025infinipotv}. A third,
complementary line reduces \emph{which} frames reach the decoder, via temporal projectors
that propagate information before pooling~\citep{jiang2025storm}, training free frame
selection~\citep{kim2026swift,tang2025aks}, or agentic coarse to fine zooming for
high resolution clips on demand~\citep{ding2026videozoomer}. All leave the retained
evidence in the activation stream, reprocessed by the full decoder for every question.
\method{} instead removes visual positions before the first decoder layer, adding a
sample specific weight space pathway that recovers the lost evidence.

\paragraph{Selective state space models for video.}
Selective state space models give linear time sequence
mixing~\citep{gu2022s4,gu2024mamba,dao2024mamba2,lahoti2026mamba3} and appear in video
encoders, hybrid decoders, and temporal enrichment
modules~\citep{li2024videomamba,ren2025vamba,jiang2025storm,liao2025mmmamba}. BIMBA
interleaves learned carrier vectors with visual embeddings and refines them with a
selective scan before decoding~\citep{islam2025bimba}. Our compressor shares this mixing
principle but adds question embeddings at both scan boundaries, preserves retained
carriers' original multimodal rotary positions, and passes the full resolution stream to a
hypernetwork after contraction. We do not claim it as a primary contribution: it is
deliberately BIMBA like, so our novelty lies in the contraction and weight correction
mechanism and in the paired analysis of what the generated update adds.

\paragraph{Generated adapters and contextual weights.}
Hypernetworks generate another network's parameters~\citep{ha2017hypernetworks}, and LoRA
makes input conditioned updates tractable at scale~\citep{hu2022lora}. Recent work
generates adapters from task descriptions, documents, edit instructions, and
prefixes~\citep{li2021prefix,mahabadi2021hyperformer,charakorn2025textlora,
charakorn2026doctolora,zeng2025hyperedit,tan2025instant,abdalla2025zhyper,
zeng2026moegen,liu2026shine}, while context distillation and fast weight methods store a
context outside the activation stream~\citep{askell2021general,snell2022context,
eyuboglu2025cartridges,wang2026learnwhat,shin2025gpi,behrouz2025titans,zhang2025lact,
ba2016fastweights,schmidhuber1992fastweights}. All of these methods condition on text.
\method{} instead generates its adapter from a compressed video forward pass and binds
generation to physical sequence contraction. Mechanically its update is a video conditioned,
hypernetwork generated LoRA in this contextual adapter
family~\citep{han2026weightsurvey,zeng2025genweights}. To our knowledge it is the first to
couple video conditioned adapter generation with physical contraction. The adapter is
produced in one amortized forward pass and compensates the frozen decoder for the removed
positions.

\section{Method}
\label{sec:method}

\method{} compresses the visual prefix that enters a frozen video language model and
reuses the removed information to generate a temporary video conditioned weight update.

\subsection{Problem setup and notation}
\label{sec:setup-method}

Let a video be a sequence of $T$ sampled frames,
$V=(I_1,\ldots,I_T)$, and let $q$ denote the text question. A frozen vision encoder
followed by the frozen multimodal projector maps frame $I_t$ to a matrix of visual
tokens,
\begin{equation}
 \mathbf X_t\in\mathbb R^{n_t\times d},\qquad t\in\{1,\ldots,T\},
 \label{eq:stream}
\end{equation}
where $d$ is the language model hidden width and $n_t$ is the number of
full resolution visual tokens for frame $t$. Bold uppercase letters denote matrices
and bold lowercase letters denote vectors. Let $n_{\mathrm{ref}}$ be the nominal
number of visual tokens per frame under the chosen visual preprocessing, and
$n_{\mathrm{vis}}=\sum_{t=1}^{T}n_t$ the total number of full resolution visual
tokens. Backbone and preprocessing are specified in the experimental protocol.

Let $\mathbf Q\in\mathbb R^{n_q\times d}$ contain the projected timestamp and
question embeddings. The original multimodal prefix is the temporally ordered
concatenation of the frame token matrices and the text embeddings,
$\mathbf X\in\mathbb R^{S\times d}$ with $S=n_{\mathrm{vis}}+n_q$. The frozen
decoder is $f_{\theta}$, with parameters $\theta$. Neither the vision encoder nor
$f_{\theta}$ is updated.

The method builds a shorter prefix $\widetilde{\mathbf X}$ and, for selected decoder
layers, a video question conditioned update $\Delta\theta(V,q)$. Its training goal
is for the frozen decoder with the short prefix and temporary update to reproduce
the behavior of the same decoder on the original prefix:
\begin{equation}
 f_{\theta+\Delta\theta(V,q)}\!\left(y\mid\widetilde{\mathbf X}\right)
 \approx f_{\theta}\!\left(y\mid\mathbf X\right).
 \label{eq:goal}
\end{equation}
This is the setup-level counterpart of the objective in
Eq.~\eqref{eq:objective}: $f_{\theta}$ realizes the conditional
$\pi_{\theta}$, and $\widetilde{\mathbf X}$ is the counterpart of the compressed
input $\mathcal{C}(X)$. We henceforth write $f_{\theta}$ for the frozen video
language model.

We distinguish two visual representations. A \emph{full resolution visual token} is
a row of $\mathbf X_t$. A \emph{carrier token} is a learned replacement for a group
of full resolution visual tokens that remains in the decoder input after
contraction.

\begin{figure}[t]
\centering
\includegraphics[width=\textwidth]{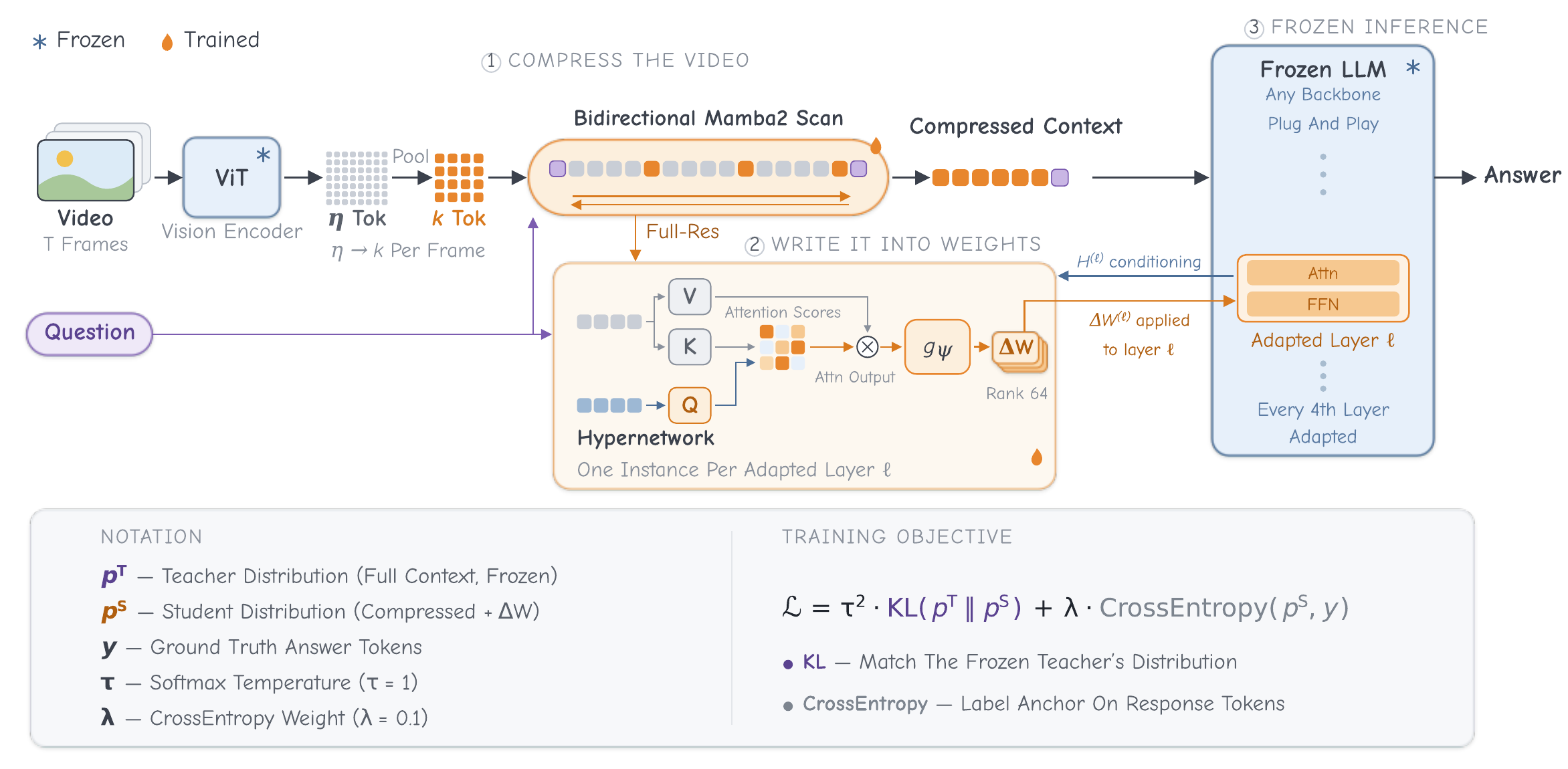}
\caption{\textbf{\method{} overview.} The frozen vision encoder produces $n_t$
full resolution visual tokens for frame $t$. Bilinear resampling initializes $k_t$
carrier tokens, and a bidirectional Mamba2 scan refines the carrier tokens while
reading the full resolution visual stream and the question. The carrier tokens
replace the original frame tokens in the decoder prefix. At selected decoder layers,
a hypernetwork uses the contracted decoder state and a full resolution per frame
readout to generate LoRA updates for selected attention and MLP projections. The
updates are merged after prefill. Decoding then uses the frozen backbone over the
contracted prefix.}
\label{fig:architecture}
\end{figure}

Figure~\ref{fig:architecture} gives an overview.

\subsection{Question conditioned visual compression}
\label{sec:compressor}

\paragraph{Carrier token initialization.}
We assign $k_t$ carrier tokens to frame $t$. Let $k$ be the target carrier token
budget for a frame with $n_{\mathrm{ref}}$ full resolution tokens, scaled to the
actual frame length as 
\begin{equation}
 k_t=\min\!\left\{n_t,\;\max\!\left\{1,\;
 \operatorname{round}\!\left(\frac{n_t}{n_{\mathrm{ref}}}k\right)\right\}\right\}.
 \label{eq:budget}
\end{equation}
The token budget and resulting compression ratio are experimental configuration
choice, reported in Section~\ref{sec:experiments}.

To preserve spatial layout at initialization, we reshape the $n_t$ tokens of a frame
to their original $h_t\times w_t$ grid ($n_t=h_tw_t$) and choose integers $a_t,b_t$
with $a_tb_t=k_t$ and $a_t/b_t$ closest to $h_t/w_t$. Bilinear resampling then
initializes the carrier token matrix,
\begin{equation}
 \mathbf C_t^{(0)}=
 \operatorname{Bilinear}\!\left(
 \operatorname{grid}_{h_t\times w_t}(\mathbf X_t),\;a_t,b_t\right)
 \in\mathbb R^{k_t\times d},
 \label{eq:bilinear}
\end{equation}
where $\operatorname{grid}_{h_t\times w_t}$ arranges the $n_t=h_tw_t$ tokens of
$\mathbf X_t$ onto their spatial grid. $\operatorname{Bilinear}$ is standard
two dimensional bilinear interpolation applied independently to each of the $d$
channels: it treats the $h_t\times w_t$ token grid as a continuous field and
resamples it onto a coarser $a_t\times b_t$ grid, so each carrier token is a convex
combination of the four full resolution tokens nearest its sampling location, with
weights given by that location's fractional distance to those tokens. Flattening the
$a_t\times b_t$ output back to a sequence gives the $k_t=a_tb_t$ carrier rows. This
initialization is parameter free. Learned compression comes only from the selective
scan below.

\paragraph{Bidirectional selective scan.}
The initial carrier tokens summarize local spatial neighborhoods but have not seen
the rest of the video or the question, so we process the full resolution tokens,
the carrier tokens, and the question jointly. Let $[\,\cdot\,;\,\cdot\,]$ denote
sequence concatenation and $\operatorname{Interleave}(\mathbf X_t,\mathbf C_t^{(0)})$
insert the $k_t$ carrier tokens approximately uniformly among the $n_t$
full resolution tokens of frame $t$. The scan input is
\begin{equation}
 \mathbf Z=\left[
 \mathbf Q;\;
 \operatorname{Interleave}(\mathbf X_1,\mathbf C_1^{(0)});\;\cdots;\;
 \operatorname{Interleave}(\mathbf X_T,\mathbf C_T^{(0)});\;
 \mathbf Q\right].
 \label{eq:scanseq}
\end{equation}
Placing $\mathbf Q$ at both ends ensures that, in at least one scan direction, every
carrier token is close to the question tokens.

A forward and a reverse Mamba2 block, with independent parameters, produce
\begin{equation}
 \mathbf Y=\mathbf Z+
 \operatorname{SSM}_{\rightarrow}\!\left(\operatorname{LN}(\mathbf Z)\right)+
 \operatorname{Reverse}\!\left[
 \operatorname{SSM}_{\leftarrow}\!\left(
 \operatorname{LN}(\operatorname{Reverse}[\mathbf Z])\right)\right].
 \label{eq:bidir}
\end{equation}
The carrier token rows of $\mathbf Y$ for frame $t$ form
$\mathbf C_t\in\mathbb R^{k_t\times d}$. The remaining visual token rows form
$\mathbf F_t\in\mathbb R^{n_t\times d}$. The carrier tokens enter the contracted
decoder prefix. The full resolution features $\mathbf F_t$ stay outside it, used only by the
hypernetwork, so contraction removes visual positions while the hypernetwork can still
consult the full resolution evidence.
\subsection{Physical sequence contraction}
\label{sec:contraction}

Carrier tokens reduce decoder cost only if the full resolution visual positions they
represent are removed. For frame $t$, partition its $n_t$ original visual positions into
$k_t$ consecutive segments of nearly equal size, and let $s_{t,j}$ be the final original
decoder position in segment $j$. We keep all text positions and only these representative
visual positions:
\begin{equation}
 \mathcal I=
 \bigl(\{1,\ldots,S\}\setminus\mathcal V\bigr)
 \;\cup\;
 \left\{s_{t,j}:1\leq t\leq T,\;1\leq j\leq k_t\right\}.
 \label{eq:keepset}
\end{equation}
Here $\mathcal V$ is the set of visual positions in the original prefix. We form
$\widetilde{\mathbf X}$ by reading the positions in $\mathcal I$ in their original order,
replacing the embedding at $s_{t,j}$ with carrier token $\mathbf C_t[j]$, and leaving every
retained text embedding unchanged. Its length is
\begin{equation}
 \widetilde S=S-n_{\mathrm{vis}}+\sum_{t=1}^{T}k_t.
 \label{eq:contracted-length}
\end{equation}

The same selection applies to the attention mask, timestamps, and multimodal rotary position
embeddings (MRoPE), so carrier token $\mathbf C_t[j]$ inherits the position of $s_{t,j}$ rather
than a new one. Instantiated input lengths and compression ratios are in
Section~\ref{sec:experiments}.

\subsection{Video conditioned LoRA updates}
\label{sec:hypernet}

The carrier tokens supply the decoder's explicit visual context. The hypernetwork adds a
temporary weight space path that can use the full resolution features $\mathbf F_t$ absent
from the contracted prefix. Let $\mathcal A$ denote the set of decoder layers with generated
updates. For each $\ell\in\mathcal A$, let $\mathbf H^{(\ell)}\in\mathbb R^{\widetilde S\times d}$
be the residual stream at the input of layer $\ell$, and let $\mathcal P_t$ be the positions
of frame $t$'s carrier tokens in the contracted prefix. The choice of $\mathcal A$ is specified
in Section~\ref{sec:experiments}.

\paragraph{Conditioning vectors.}
The hypernetwork receives three conditioning vectors from each adapted layer: a carrier
summary, a full resolution readout, and the final question state. The first averages the hidden
states at every carrier token position. With $N_C=\sum_{t=1}^{T}k_t$ and
$\mathcal P=\bigcup_t\mathcal P_t$,
\begin{equation}
 \mathbf v^{(\ell)}=
 \frac{1}{N_C}\sum_{i\in\mathcal P}\mathbf H^{(\ell)}[i]
 \in\mathbb R^d.
 \label{eq:visual-summary}
\end{equation}

Second, each carrier token state attends to the full resolution features of its own frame.
Let $\mathbf R_t^{(\ell)}=\mathbf H^{(\ell)}[\mathcal P_t]$ be the $k_t\times d$ matrix of
carrier token states for frame $t$. Using a shared multihead cross attention block with
residual connection,
\begin{equation}
 \overline{\mathbf R}_t^{(\ell)}=
 \mathbf R_t^{(\ell)}+
 \operatorname{MHA}\!\left(
 \operatorname{LN}(\mathbf R_t^{(\ell)}),\;
 \mathbf F_t,\;\mathbf F_t\right),
 \qquad
 \mathbf c^{(\ell)}=
 \frac{1}{N_C}\sum_{t=1}^{T}\sum_{j=1}^{k_t}
 \overline{\mathbf R}_t^{(\ell)}[j].
 \label{eq:crosspool}
\end{equation}
The $\operatorname{MHA}$ projections are trainable, and attention is independent across frames
(frame $t$'s carriers attend only to $\mathbf F_t$). The vector $\mathbf c^{(\ell)}\in\mathbb R^d$
is the mean full resolution readout across carrier tokens, and the factor $1/N_C$ keeps its
scale independent of frame and carrier count.

Third, $\mathbf q^{(\ell)}=\mathbf H^{(\ell)}[p]$ is the state at the final question token
position $p$, which attends to the preceding video and question under the causal mask. We
concatenate the three,
\begin{equation}
 \mathbf g^{(\ell)}=
 \left[\mathbf v^{(\ell)};\mathbf c^{(\ell)};\mathbf q^{(\ell)}\right]
 \in\mathbb R^{3d}.
 \label{eq:conditioning}
\end{equation}

\paragraph{Generating the update.}
The hypernetwork maps $\mathbf g^{(\ell)}$ to a latent vector
$\mathbf z^{(\ell)}\in\mathbb R^{d_z}$. At each adapted layer, let $m$ index a selected decoder
projection with frozen weight
$\mathbf W_m^{(\ell)}\in\mathbb R^{d_{\mathrm{out},m}\times d_{\mathrm{in},m}}$. For a chosen
LoRA rank $r$, a module specific head $\mathbf P_m^{(\ell)}\in\mathbb R^{r d_z\times d_z}$
produces a normalized latent matrix, which two module specific projections
$\mathbf U_m^{(\ell)}\in\mathbb R^{d_z\times d_{\mathrm{in},m}}$ and
$\mathbf V_m^{(\ell)}\in\mathbb R^{d_z\times d_{\mathrm{out},m}}$ map to the two LoRA factors:
\begin{equation}
 \mathbf Z_m^{(\ell)}=
 \operatorname{RowNorm}\!\left(
 \operatorname{reshape}_{r\times d_z}
 \left(\mathbf P_m^{(\ell)}\mathbf z^{(\ell)}\right)\right),
 \qquad
 \mathbf A_m^{(\ell)}=\mathbf Z_m^{(\ell)}\mathbf U_m^{(\ell)},
 \qquad
 \mathbf B_m^{(\ell)}=\mathbf Z_m^{(\ell)}\mathbf V_m^{(\ell)}.
 \label{eq:lora-factors}
\end{equation}
Here $\operatorname{RowNorm}$ normalizes each row to unit $\ell_2$ norm, so
$\mathbf A_m^{(\ell)}\in\mathbb R^{r\times d_{\mathrm{in},m}}$ and
$\mathbf B_m^{(\ell)}\in\mathbb R^{r\times d_{\mathrm{out},m}}$, and the correction matches the
frozen projection's shape:
\begin{equation}
 \Delta\mathbf W_m^{(\ell)}=
 \frac{\alpha}{r}\left(\mathbf B_m^{(\ell)}\right)^\top
 \mathbf A_m^{(\ell)},
 \label{eq:lora}
\end{equation}
with $\alpha$ the LoRA scale. Initializing each $\mathbf V_m^{(\ell)}$ to zero makes every
$\Delta\mathbf W_m^{(\ell)}$ zero at initialization, so training begins from the frozen decoder
on the contracted prefix. Architecture, rank, scale, adapted projections, and parameter count
are in Section~\ref{sec:experiments}.

\paragraph{Prefill and decoding.}
During prefill, we apply the correction without forming a dense
$d_{\mathrm{out},m}\times d_{\mathrm{in},m}$ matrix. For an input matrix
$\mathbf X_{\mathrm{in}}$, the modified projection is
\begin{equation}
 \operatorname{Linear}^{(\ell)}_m(\mathbf X_{\mathrm{in}})=
 \mathbf X_{\mathrm{in}}\left(\mathbf W_m^{(\ell)}\right)^\top+
 \frac{\alpha}{r}
 \left(\mathbf X_{\mathrm{in}}\left(\mathbf A_m^{(\ell)}\right)^\top\right)
 \mathbf B_m^{(\ell)}.
 \label{eq:functional}
\end{equation}
The update depends only on the video question prefix, so it is constant across answer tokens.
After prefill we merge the corrections into the frozen projections, decode with the ordinary
frozen decoder over the shorter KV cache, and subtract them before the next example. The
hypernetwork and low rank multiplication are never evaluated per generated token.

\subsection{Training by context distillation}
\label{sec:training}

We train the compressor and hypernetworks by distilling from the frozen
backbone~\citep{hinton2015distilling} on the uncompressed prefix. For each example the teacher
computes logits on $\mathbf X$ without gradients and the student on $\widetilde{\mathbf X}$ with
its generated updates (response positions are never contracted). With $\mathcal R$ the
response token positions, $\mathbf u_r^{\mathrm T},\mathbf u_r^{\mathrm S}$ the teacher and
student logits at $r$, and $y_r$ the answer token, define
\begin{equation}
 \mathbf p_r^{\mathrm T}=\operatorname{softmax}
 \left(\frac{\mathbf u_r^{\mathrm T}}{\tau}\right),
 \qquad
 \mathbf p_r^{\mathrm S}=\operatorname{softmax}
 \left(\frac{\mathbf u_r^{\mathrm S}}{\tau}\right).
 \label{eq:teacher-student}
\end{equation}
We minimize
\begin{equation}
 \mathcal L=
 \frac{1}{|\mathcal R|}\sum_{r\in\mathcal R}
 \left[
 \tau^2\operatorname{KL}\!\left(
 \mathbf p_r^{\mathrm T}\,\middle\|\,\mathbf p_r^{\mathrm S}\right)
 +\lambda\operatorname{CE}\!\left(\mathbf u_r^{\mathrm S},y_r\right)
 \right].
 \label{eq:distillation}
\end{equation}
The KL term transfers the frozen model's full vocabulary distribution, and the cross entropy
term anchors the student to the observed response.
\section{Experiments}
\label{sec:experiments}

\paragraph{Architecture configuration.}
All experiments use Qwen3-VL-8B-Instruct as the frozen backbone (36 decoder layers of
width 4,096). Neither its vision encoder nor its decoder is updated, and the fixed
visual preprocessing produces $n_{\mathrm{ref}}=56$ full resolution visual tokens per
frame. The compressor is one forward and one reverse Mamba2 block with state size 64.
The hypernetwork uses a two layer trunk with latent width $d_z=256$, adapts layers
$\mathcal A=\{3,7,\ldots,35\}$ (every fourth layer), and generates rank-$r=64$ updates
with scale $\alpha=128$ for the query, key, value, output, gate, up, and down
projections. The trainable system totals 0.98B parameters (763M hypernetwork, 203M
compressor, 16.8M shared cross attention).

\paragraph{Carrier token budgets and training.}
Unless stated otherwise, the primary model uses $k=16$ carrier tokens per frame,
compared with $n_{\mathrm{ref}}=56$ full resolution visual tokens per frame. We
also train models with $k\in\{12,8,6\}$ to trace the accuracy versus efficiency frontier.
We train on 1.65M multiple choice and open ended video question answering examples
from LLaVA-Video-178K~\citep{zhang2024llavavideo}, with 128 sampled frames per
training video. AdamW uses a peak learning rate of $3\times10^{-5}$, 3\% warmup,
cosine decay, bfloat16 arithmetic, and effective batch size 120. The distillation
loss uses $\tau=1$ and $\lambda=0.1$.

\paragraph{Evaluation and paired update analysis.}
All efficient models are evaluated at 256 uniformly sampled frames with greedy
decoding on NextQA, PerceptionTest, MVBench, TOMATO, Video-MME with and without
subtitles, LongVideoBench, MLVU, and LVBench~\citep{xiao2021nextqa,patraucean2023perceptiontest,li2024mvbench,shangguan2025tomato,fu2025videomme,wu2024longvideobench,zhou2025mlvu,wang2025lvbench}.
\method{} trains at 128 frames but no component is retrained or rescaled for longer clips. The
carrier budget scaling (Eq.~\eqref{eq:budget}) and inherited multimodal rotary position
embeddings (MRoPE) are frame count agnostic. We rerun BIMBA, Vamba, and VideoNSA under the same
protocol within their supported frame ranges. Latency is on one B200 GPU at batch size one in
bfloat16, with the full grid in Table~\ref{tab:latency-grid}.
\section{Results}
\label{sec:results}

\subsection{\method{} provides the strongest complete coverage efficient result}
\label{sec:results-main}


\begin{table*}[!t]
\caption{\textbf{Video question answering accuracy}, in percent. The short
  average covers the first four columns and the long average the next five.
  Rows labelled \method{} give our compressor and hypernetwork on a frozen
  Qwen3-VL-8B-Instruct backbone at four values of the per frame token budget
  $k$, all at $256$ sampled frames. Bold marks the best entry among the
  efficient models re-evaluated under our controlled $256$-frame protocol
  (\method{}, BIMBA, Vamba, VideoNSA). MLVU and LVBench entries are the
  \emph{test} split. Table~\ref{tab:mainfull} in the appendix adds proprietary
  models, additional open baselines, and further efficient methods with the
  accuracies stated in their own papers.}
\label{tab:main}
\centering
\setlength{\tabcolsep}{3pt}
\renewcommand{\arraystretch}{0.78}
\resizebox{\textwidth}{!}{%
\begin{tabular}{@{}lcccccccccccc@{}}
\toprule
\textbf{Model}
 & \benchcell{NextQA}{test \citep{xiao2021nextqa}}
 & \benchcell{PerceptionTest}{test \citep{patraucean2023perceptiontest}}
 & \benchcell{MVBench}{test \citep{li2024mvbench}}
 & \benchcell{Tomato}{test \citep{shangguan2025tomato}}
 & \benchcell{Short QA avg.}{}
 & \benchcell{Video-MME}{test \citep{fu2025videomme}}
 & \benchcell{Video-MME-Sub}{test \citep{fu2025videomme}}
 & \benchcell{LongVideoBench}{val \citep{wu2024longvideobench}}
 & \benchcell{MLVU}{test MCQ \citep{zhou2025mlvu}}
 & \benchcell{LVBench}{test \citep{wang2025lvbench}}
 & \benchcell{Long QA avg.}{}
 & \benchcell{Average}{}\\
\midrule
\multicolumn{13}{@{}l}{\emph{Open models}}\\
Qwen3-VL-8B \citep{bai2025qwen3vl}
 & 83.4 & 72.7 & 68.7 & 35.7 & 65.1
 & 71.4 & 75.2 & 62.4 & 57.6 & 58.0 & 64.9
 & 65.0\\
Eagle2.5-8B \citep{chen2025eagle25}
 & 85.0 & 81.0 & 74.8 & 31.0 & 68.0
 & 72.4 & 75.7 & 66.4 & 60.4 & 50.9 & 65.2
 & 66.4\\
PLM-8B \citep{cho2025perceptionlm}
 & 84.1 & 82.7 & 77.1 & 33.2 & 69.3
 & 58.3 & 65.4 & 56.9 & 52.6 & 44.5 & 55.5
 & 61.6\\
LLaVA-Video-7B \citep{zhang2024llavavideo}
 & 83.2 & 68.8 & 58.6 & 24.9 & 58.9
 & 63.3 & 69.7 & 58.2 & 52.8 & 44.2 & 57.6
 & 58.2\\
Molmo2-8B \citep{molmo2team2026}
 & 86.2 & 82.1 & 75.9 & 39.6 & 71.0
 & 69.9 & 75.8 & 67.5 & 60.2 & 52.8 & 65.2
 & 67.8\\
\midrule
\multicolumn{13}{@{}l}{\emph{Efficient video language models} (controlled re-evaluation at 256 frames)}\\
BIMBA \citep{islam2025bimba}
 & 80.9 & 65.8 & 59.6 & 24.0 & 57.6
 & 63.3 & 66.4 & 57.7 & \textbf{53.9} & 43.3 & 56.9
 & 57.2\\
Vamba \citep{ren2025vamba}
 & 77.5 & \textbf{69.4} & 59.6 & 26.2 & 58.2
 & 58.1 & 61.2 & 52.8 & 45.7 & 41.2 & 51.8
 & 54.6\\
VideoNSA \citep{song2025videonsa}
 & 77.1 & 56.7 & 57.0 & 22.7 & 53.4
 & 59.4 & 61.9 & 52.2 & 41.3 & 41.8 & 51.3
 & 52.2\\
\rowcolor{lightred}
\method{} ($k = 16$)
 & \textbf{81.8} & 66.7 & \textbf{62.8} & \textbf{30.9} & \textbf{60.6}
 & \textbf{66.2} & \textbf{67.5} & \textbf{60.4} & 52.0 & \textbf{44.0} & \textbf{58.0}
 & \textbf{59.1}\\
\method{} ($k = 12$)
 & 79.6 & 66.2 & 60.9 & 29.3 & 59.0
 & 64.1 & 66.5 & 59.6 & 52.5 & 42.0 & 56.9
 & 57.9\\
\method{} ($k = 8$)
 & 80.0 & 65.4 & 60.6 & 28.7 & 58.7
 & 63.8 & 66.3 & 58.8 & 51.9 & 40.0 & 56.2
 & 57.3\\
\method{} ($k = 6$)
 & 79.9 & 63.4 & 60.0 & 28.4 & 57.9
 & 62.0 & 65.5 & 58.6 & 50.1 & 41.3 & 55.5
 & 56.6\\
\bottomrule
\end{tabular}%
}
\end{table*}

Table~\ref{tab:main} compares \method{} against efficient video llms we
reevaluate under a common 256 frame budget. We match frame count, decoding, hardware,
and evaluation suite, while backbone, training data, and architecture stay method
specific. This is a matched evaluation budget, not a controlled architectural comparison.
Among these rows, \method{} with $k=16$ carrier tokens attains the highest short, long,
and overall averages, leading the short form, multimodal, subtitle aware, and long video
evaluations at once. It stays competitive down to $k=6$. It trails only on two
columns: Vamba on PerceptionTest, whose recurrent cross attention keeps every video token,
and BIMBA on MLVU.

\begin{figure*}[t]
\centering
\includegraphics[width=0.93\textwidth]{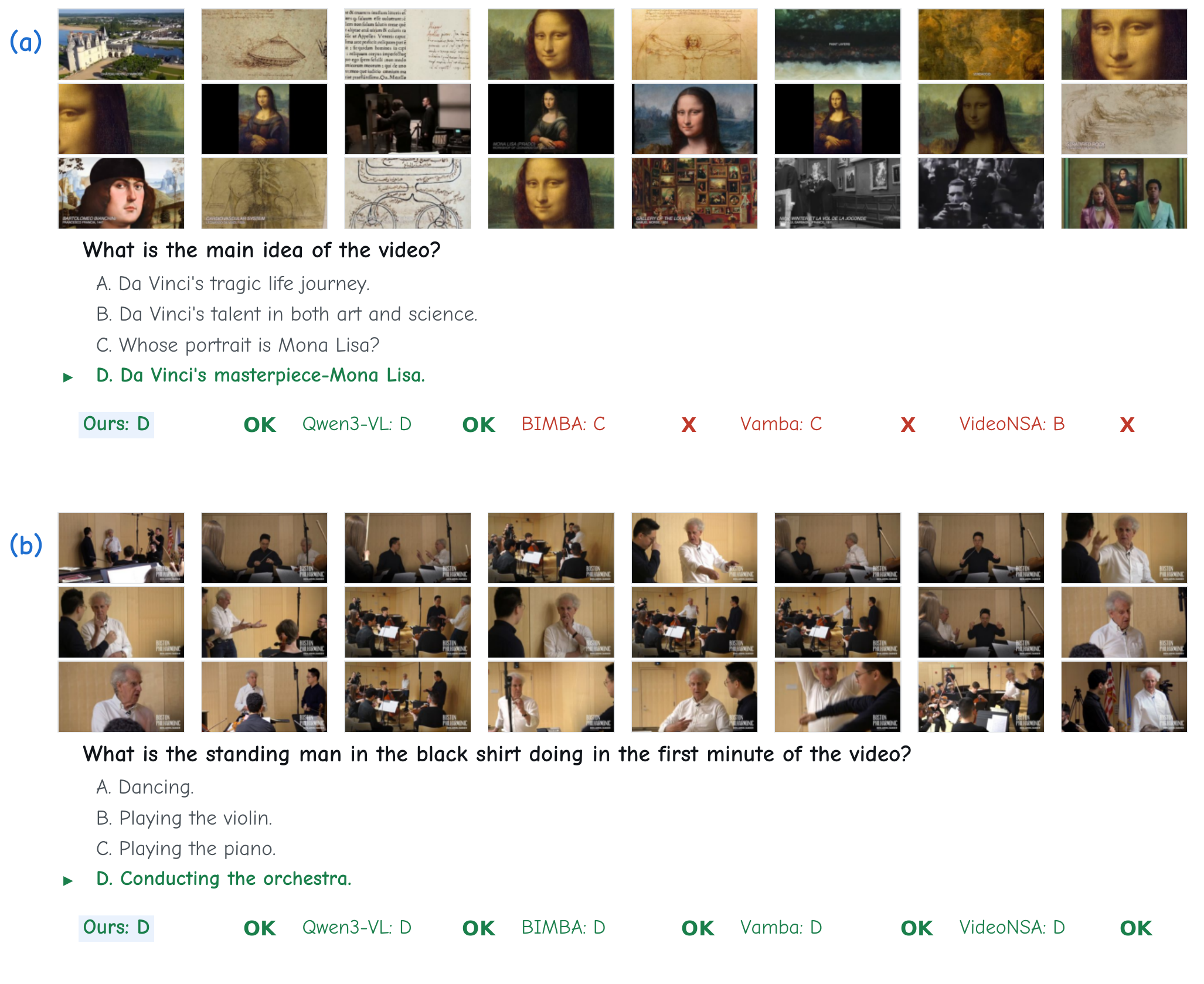}
\caption{\textbf{Qualitative examples at 256 frames.} On a question about a video's global
subject, \method{} matches the uncompressed backbone while the efficient baselines pick
locally plausible distractors. On a temporally localized question all systems answer
correctly.}
\label{fig:qualitative}
\end{figure*}

Table~\ref{tab:mainfull} lists further efficient models at their reported
accuracies. With different backbones, and protocols they are not directly
comparable, so we neither bold nor aggregate them, but on the benchmarks each reports,
\method{} is competitive with or ahead of them. These methods span the design space around
\method{}~\citep{ren2025vamba,song2025videonsa,shen2024longvu,zohar2025apollo,tao2025dycoke,
jiang2025storm,ding2026videozoomer,kim2026swift} but all keep their evidence in the activation
stream, whereas \method{} contracts it and moves the missing evidence into a weight update.
\method{} trades part of that visual context for a much shorter decoder sequence, accuracy
declining smoothly from $k=16$ to $k=6$. Appendix~\ref{app:baselines} details every baseline
and Figure~\ref{fig:qualitative} gives examples.

\subsection{The generated update does task-relevant work}
\label{sec:results-weight}

\begin{table}[t]
\centering
\footnotesize
\setlength{\tabcolsep}{6pt}
\caption{\textbf{Paired update analysis at fixed compressed context.} Disc.\ discards the
per example LoRA update and Appl.\ applies it; $\Delta$ is the paired difference in points at
identical carrier context. $^{\dagger}$ marks subsets not significant under the exact McNemar test.}
\label{tab:weight-evidence}
\begin{tabular}{@{}lrrr@{\hspace{3em}}lrrr@{}}
\toprule
Subset & Disc. & Appl. & $\Delta$ & Subset & Disc. & Appl. & $\Delta$ \\
\cmidrule(r{1.5em}){1-4}\cmidrule{5-8}
\multicolumn{4}{@{}l}{\emph{Video-MME}} & \multicolumn{4}{l@{}}{\emph{LongVideoBench}} \\
\quad Counting        & 40.67 & 44.03 & $+3.36$ & \quad Sub.\,$\to$\,next               & 45.21 & 57.53 & $+12.32$ \\
\quad Law             & 61.19 & 67.62 & $+6.43$ & \quad Scene\,$\to$\,sub.$^{\dagger}$  & 32.88 & 43.84 & $+10.96$ \\
\quad Athletics       & 54.36 & 59.23 & $+4.87$ & \quad Sub.\,$\to$\,object$^{\dagger}$ & 55.41 & 66.22 & $+10.81$ \\
\multicolumn{4}{@{}l}{\emph{MVBench}} & \multicolumn{4}{l@{}}{\emph{MLVU}} \\
\quad Action count    & 46.33 & 51.83 & $+5.50$ & \quad Count & 28.33 & 33.33 & $+5.00$ \\
\quad Counterfactual  & 51.67 & 56.33 & $+4.66$ & & & & \\
\bottomrule
\end{tabular}
\end{table}

The paired update analysis in Table~\ref{tab:weight-evidence} isolates the update's
  contribution: each checkpoint is run twice, toggling only the generated update while the
  video, question, carrier tokens, prompt, and decoding are held fixed (exact protocol in
  Appendix~\ref{app:implementation}). Any accuracy difference is therefore attributable to
  the generated update alone. Applying it raises whole benchmark accuracy by $+3.96$ on
  Video-MME, $+3.22$ on MVBench, and $+4.39$ on LongVideoBench. The subset breakdown,
  organized \emph{a priori} around the capabilities that compression should strain, shows
  the largest gains on exactly those subsets (counting, temporal action, counterfactual
  inference, and subtitle grounding), reaching $+12.32$ points on LongVideoBench subtitle
  grounding. The update is most valuable precisely where the carrier tokens lose the most,
  indicating that it restores evidence removed by contraction rather than applying a
  generic answer calibration.

\subsection{Compressor ablations}
\label{sec:compressor-ablation}

We ablate the  conditioned bidirectional selective scan of Section~\ref{sec:compressor},
holding \method{} fixed at $k{=}16$ on Video-MME without subtitles (full compressor $66.2$,
Table~\ref{tab:compressor-ablation}). Each variant changes only the compressor, so the
differences isolate how the carrier tokens are formed. The largest effect
comes from question conditioning: removing the question from the scan input at test time drops
accuracy to $64.3$ ($-1.9$). At $k{=}16$ each carrier token stands in for $3.5$ full resolution
tokens, so the compressor cannot keep everything; the drop shows it exploits the question to
decide what the carrier tokens retain. A retrained forward only scan reaches $65.1$ ($-1.1$):
with one direction, early carrier tokens never see later content or the trailing question,
whereas the reverse scan with $\mathbf Q$ at both ends (Eq.~\eqref{eq:scanseq}) puts every
carrier token near the question in at least one direction. Replacing the scan with bilinear
pooling alone (Eq.~\eqref{eq:bilinear}) gives $64.7$ ($-1.5$): the parameter free
initialization is a strong baseline but purely local, each carrier token a fixed combination
of adjacent tokens in its own frame, so the recovered $+1.5$ is the value of learned, question
aware mixing across the whole sequence. A needle in a haystack probe of what the compressed
  context preserves across carrier budgets is given in Appendix~\ref{sec:vniah}.

\begin{table}[t]
\centering
\small
\caption{Compressor ablations at $k{=}16$ on Video-MME without subtitles (accuracy \%,
$\Delta$ vs.\ the full compressor). Each variant changes only the compressor.}
\label{tab:compressor-ablation}
\begin{tabularx}{\linewidth}{@{}Xcc@{}}
\toprule
Compressor variant & Video-MME (no sub.) & $\Delta$ \\
\midrule
\rowcolor{lightred}
\method{} (full compressor)                          & 66.2 & --- \\
\quad $-$ question conditioning (test time)           & 64.3 & $-1.9$ \\
\quad $-$ bidirectional scan (forward only, retrained) & 65.1 & $-1.1$ \\
\quad $-$ selective scan (bilinear pooling only, retrained) & 64.7 & $-1.5$ \\
\bottomrule
\end{tabularx}
\end{table}

\section{Discussion}
\label{sec:conclusion}

\method{} splits video-conditioned computation between a short explicit carrier-token context and a temporary, video-conditioned weight update generated by hypernetworks and merged after prefill. At 16 carrier tokens per frame, it contracts the visual stream by $3.5\times$, reducing decoder prefill time by $3.35\times$ at 256 frames while maintaining near-constant autoregressive decoding latency and proportionally reducing KV-cache usage. Under a matched 256-frame budget, \method{} achieves the strongest average accuracy among the efficient video models evaluated across nine benchmark settings. Moreover, paired ablations that hold the compressed visual context fixed show that the generated weight update performs task-relevant computation beyond the information retained by the carrier tokens, with gains of up to 4.4 points on established video understanding benchmarks.

\paragraph{Limitations}
\label{sec:limitations}
Several claims are scoped by present evidence. The paired update does task relevant work, but
its video specificity remains a hypothesis awaiting the mismatched video control of
Appendix~\ref{app:implementation}. \method{} uses a single frozen backbone (Qwen3-VL-8B) and
public benchmarks, so its generality is untested; the hypernetwork adds prefill compute, so
savings appear only when generation is long relative to the video; and the main comparisons
report a matched evaluation budget rather than a controlled architectural comparison.

\section*{Ethics statement}
\method{} builds on a frozen public backbone and public video question answering
benchmarks and introduces no new data collection or human subjects. It inherits the
biases and failure modes of its backbone. Cheaper long video understanding can aid
accessibility and analysis but can also lower the cost of harmful large scale video
surveillance or profiling; we intend the method for research and beneficial use.

\section*{Reproducibility statement}
Section~\ref{sec:method} gives the full method and objective, Section~\ref{sec:experiments}
lists the backbone, carrier budgets, adapted layers, ranks, optimizer, and training data,
and Appendix~\ref{app:implementation} details the selective scan recurrence, adapter head
shapes, DeepStack alignment, two pass training, and the exact paired update and latency
protocols. All benchmarks and backbones are public, so the reported results can be
reproduced from these specifications.

\section*{Use of AI assistants}
AI based tools were used only to polish grammar, wording, and phrasing of the manuscript.
They were not used to generate research ideas, design or run experiments, produce results,
or draw conclusions, all of which are the authors' own work.

\bibliographystyle{iclr2027_conference}
\bibliography{iclr2027_conference}

@inproceedings{ha2017hypernetworks,
  title     = {HyperNetworks},
  author    = {Ha, David and Dai, Andrew M. and Le, Quoc V.},
  booktitle = {International Conference on Learning Representations (ICLR)},
  year      = {2017},
  note      = {arXiv:1609.09106}
}

@inproceedings{
hu2022lora,
title={Lo{RA}: Low-Rank Adaptation of Large Language Models},
author={Edward J Hu and yelong shen and Phillip Wallis and Zeyuan Allen-Zhu and Yuanzhi Li and Shean Wang and Lu Wang and Weizhu Chen},
booktitle={International Conference on Learning Representations},
year={2022},
url={https://openreview.net/forum?id=nZeVKeeFYf9}
}

@article{hinton2015distilling,
  title={Distilling the knowledge in a neural network},
  author={Hinton, Geoffrey and Vinyals, Oriol and Dean, Jeff},
  journal={arXiv preprint arXiv:1503.02531},
  year={2015}
}

@article{askell2021general,
  title={A general language assistant as a laboratory for alignment},
  author={Askell, Amanda and Bai, Yuntao and Chen, Anna and Drain, Dawn and Ganguli, Deep and Henighan, Tom and Jones, Andy and Joseph, Nicholas and Mann, Ben and DasSarma, Nova and others},
  journal={arXiv preprint arXiv:2112.00861},
  year={2021}
}

@article{snell2022context,
  title={Learning by distilling context},
  author={Snell, Charlie and Klein, Dan and Zhong, Ruiqi},
  journal={arXiv preprint arXiv:2209.15189},
  year={2022}
}

@article{mu2023gisting,
  title={Learning to compress prompts with gist tokens},
  author={Mu, Jesse and Li, Xiang and Goodman, Noah},
  journal={Advances in Neural Information Processing Systems},
  volume={36},
  pages={19327--19352},
  year={2023}
}

@inproceedings{
gu2024mamba,
title={Mamba: Linear-Time Sequence Modeling with Selective State Spaces},
author={Albert Gu and Tri Dao},
booktitle={First Conference on Language Modeling},
year={2024},
}

@article{dao2024mamba2,
  title={Transformers are SSMs: Generalized models and efficient algorithms through structured state space duality},
  author={Dao, Tri and Gu, Albert},
  booktitle={International Conference on Machine Learning (ICML)},
  year={2024}
}

@inproceedings{li2024videomamba,
author = {Li, Kunchang and Li, Xinhao and Wang, Yi and He, Yinan and Wang, Yali and Wang, Limin and Qiao, Yu},
title = {VideoMamba: State Space Model for Efficient Video Understanding},
year = {2024},
booktitle = {Computer Vision – ECCV 2024: 18th European Conference, Milan, Italy, September 29–October 4, 2024, Proceedings, Part XXVI},
}

@misc{bai2025qwen25vl,
      title={Qwen2.5-VL Technical Report}, 
      author={Shuai Bai and Keqin Chen and Xuejing Liu and Jialin Wang and Wenbin Ge and Sibo Song and Kai Dang and Peng Wang and Shijie Wang and Jun Tang and Humen Zhong and Yuanzhi Zhu and Mingkun Yang and Zhaohai Li and Jianqiang Wan and Pengfei Wang and Wei Ding and Zheren Fu and Yiheng Xu and Jiabo Ye and Xi Zhang and Tianbao Xie and Zesen Cheng and Hang Zhang and Zhibo Yang and Haiyang Xu and Junyang Lin},
      year={2025},
      eprint={2502.13923},
      archivePrefix={arXiv},
      primaryClass={cs.CV},
      url={https://arxiv.org/abs/2502.13923}, 
}

@misc{bai2025qwen3vl,
      title={Qwen3-VL Technical Report}, 
      author={Shuai Bai and Yuxuan Cai and Ruizhe Chen and Keqin Chen and Xionghui Chen and Zesen Cheng and Lianghao Deng and Wei Ding and Chang Gao and Chunjiang Ge and Wenbin Ge and Zhifang Guo and Qidong Huang and Jie Huang and Fei Huang and Binyuan Hui and Shutong Jiang and Zhaohai Li and Mingsheng Li and Mei Li and Kaixin Li and Zicheng Lin and Junyang Lin and Xuejing Liu and Jiawei Liu and Chenglong Liu and Yang Liu and Dayiheng Liu and Shixuan Liu and Dunjie Lu and Ruilin Luo and Chenxu Lv and Rui Men and Lingchen Meng and Xuancheng Ren and Xingzhang Ren and Sibo Song and Yuchong Sun and Jun Tang and Jianhong Tu and Jianqiang Wan and Peng Wang and Pengfei Wang and Qiuyue Wang and Yuxuan Wang and Tianbao Xie and Yiheng Xu and Haiyang Xu and Jin Xu and Zhibo Yang and Mingkun Yang and Jianxin Yang and An Yang and Bowen Yu and Fei Zhang and Hang Zhang and Xi Zhang and Bo Zheng and Humen Zhong and Jingren Zhou and Fan Zhou and Jing Zhou and Yuanzhi Zhu and Ke Zhu},
      year={2025},
      eprint={2511.21631},
      archivePrefix={arXiv},
      primaryClass={cs.CV},
      url={https://arxiv.org/abs/2511.21631}, 
}

@misc{zhang2024llavavideo,
      title={LLaVA-Video: Video Instruction Tuning With Synthetic Data}, 
      author={Yuanhan Zhang and Jinming Wu and Wei Li and Bo Li and Zejun Ma and Ziwei Liu and Chunyuan Li},
      year={2025},
      eprint={2410.02713},
      archivePrefix={arXiv},
      primaryClass={cs.CV},
      url={https://arxiv.org/abs/2410.02713}, 
}

@misc{fu2025videomme,
      title={Video-MME: The First-Ever Comprehensive Evaluation Benchmark of Multi-modal LLMs in Video Analysis}, 
      author={Chaoyou Fu and Yuhan Dai and Yongdong Luo and Lei Li and Shuhuai Ren and Renrui Zhang and Zihan Wang and Chenyu Zhou and Yunhang Shen and Mengdan Zhang and Peixian Chen and Yanwei Li and Shaohui Lin and Sirui Zhao and Ke Li and Tong Xu and Xiawu Zheng and Enhong Chen and Caifeng Shan and Ran He and Xing Sun},
      year={2025},
      eprint={2405.21075},
      archivePrefix={arXiv},
      primaryClass={cs.CV},
      url={https://arxiv.org/abs/2405.21075}, 
}

@misc{wu2024longvideobench,
      title={LongVideoBench: A Benchmark for Long-context Interleaved Video-Language Understanding}, 
      author={Haoning Wu and Dongxu Li and Bei Chen and Junnan Li},
      year={2024},
      eprint={2407.15754},
      archivePrefix={arXiv},
      primaryClass={cs.CV},
      url={https://arxiv.org/abs/2407.15754}, 
}

@misc{zhou2025mlvu,
      title={MLVU: Benchmarking Multi-task Long Video Understanding}, 
      author={Junjie Zhou and Yan Shu and Bo Zhao and Boya Wu and Zhengyang Liang and Shitao Xiao and Minghao Qin and Xi Yang and Yongping Xiong and Bo Zhang and Tiejun Huang and Zheng Liu},
      year={2025},
      eprint={2406.04264},
      archivePrefix={arXiv},
      primaryClass={cs.CV},
      url={https://arxiv.org/abs/2406.04264}, 
}

@inproceedings{chen2024fastv,
author = {Chen, Liang and Zhao, Haozhe and Liu, Tianyu and Bai, Shuai and Lin, Junyang and Zhou, Chang and Chang, Baobao},
title = {An Image is Worth 1/2 Tokens After Layer 2: Plug-and-Play Inference Acceleration for Large Vision-Language Models},
year = {2024},
booktitle = {Computer Vision – ECCV 2024: 18th European Conference, Milan, Italy, September 29–October 4, 2024, Proceedings, Part LXXXI},

}

@article{yang2025visionzip,
  title={VisionZip: Longer is Better but Not Necessary in Vision Language Models},
  author={Yang, Senqiao and Chen, Yukang and Tian, Zhuotao and Wang, Chengyao and Li, Jingyao and Yu, Bei and Jia, Jiaya},
  journal={arXiv preprint arXiv:2412.04467},
  year={2024}
}

@inproceedings{huang2025prunevid,
    title = "{P}rune{V}id: Visual Token Pruning for Efficient Video Large Language Models",
    author = "Huang, Xiaohu  and
      Zhou, Hao  and
      Han, Kai",
    booktitle = "Findings of the Association for Computational Linguistics: ACL 2025",
    month = jul,
    year = "2025",
    address = "Vienna, Austria",
    publisher = "Association for Computational Linguistics",
}

@inproceedings{tao2025dycoke,
  title={DyCoke: Dynamic Compression of Tokens for Fast Video Large Language Models},
  author={Tao, Keda and Qin, Can and You, Haoxuan and Sui, Yang and Wang, Huan},
  booktitle={Proceedings of the Computer Vision and Pattern Recognition Conference},
  pages={18992--19001},
  year={2025}
}

@InProceedings{ye2025vocollama,
    author    = {Ye, Xubing and Gan, Yukang and Huang, Xiaoke and Ge, Yixiao and Tang, Yansong},
    title     = {VoCo-LLaMA: Towards Vision Compression with Large Language Models},
    booktitle = {Proceedings of the Computer Vision and Pattern Recognition Conference (CVPR)},
    month     = {June},
    year      = {2025},
    pages     = {29836-29846}
}

@inproceedings{
shen2024longvu,
title={Long{VU}: Spatiotemporal Adaptive Compression for Long Video-Language Understanding},
author={Xiaoqian Shen and Yunyang Xiong and Changsheng Zhao and Lemeng Wu and Jun Chen and Chenchen Zhu and Zechun Liu and Fanyi Xiao and Balakrishnan Varadarajan and Florian Bordes and Zhuang Liu and Hu Xu and Hyunwoo J. Kim and Bilge Soran and Raghuraman Krishnamoorthi and Mohamed Elhoseiny and Vikas Chandra},
booktitle={Forty-second International Conference on Machine Learning},
year={2025},
}

@article{tang2025aks,
  title={Adaptive Keyframe Sampling for Long Video Understanding},
  author={Tang, Xi and Qiu, Jihao and Xie, Lingxi and Tian, Yunjie and Jiao, Jianbin and Ye, Qixiang},
  journal={arXiv preprint arXiv:2502.21271},
  year={2025}
}

@article{cai2025matryoshka,
  title={Matryoshka Multimodal Models},
  author={Cai, Mu and Yang, Jianwei and Gao, Jianfeng and Lee, Yong Jae},
  journal={Proceedings of the International Conference on Learning Representation},
  year={2025}
}

@inproceedings{
tu2025vlcache,
title={{VL}-Cache: Sparsity and Modality-Aware {KV} Cache Compression for Vision-Language Model Inference Acceleration},
author={Dezhan Tu and Danylo Vashchilenko and Yuzhe Lu and Panpan Xu},
booktitle={The Thirteenth International Conference on Learning Representations},
year={2025},
url={https://openreview.net/forum?id=HMrcv7Q4Ub}
}

@misc{qin2025videoxl2,
      title={Video-XL-2: Towards Very Long-Video Understanding Through Task-Aware KV Sparsification}, 
      author={Minghao Qin and Xiangrui Liu and Zhengyang Liang and Yan Shu and Huaying Yuan and Juenjie Zhou and Shitao Xiao and Bo Zhao and Zheng Liu},
      year={2025},
      eprint={2506.19225},
      archivePrefix={arXiv},
      primaryClass={cs.CV},
      url={https://arxiv.org/abs/2506.19225}, 
}

@inproceedings{lin2025vtw,
  title={Boosting multimodal large language models with visual tokens withdrawal for rapid inference},
  author={Lin, Zhihang and Lin, Mingbao and Lin, Luxi and Ji, Rongrong},
  booktitle={Proceedings of the AAAI Conference on Artificial Intelligence},
  volume={39},
  number={5},
  pages={5334--5342},
  year={2025}
}

@inproceedings{zohar2025apollo,
  title={Apollo: An exploration of video understanding in large multimodal models},
  author={Zohar, Orr and Wang, Xiaohan and Dubois, Yann and Mehta, Nikhil and Xiao, Tong and Hansen-Estruch, Philippe and Yu, Licheng and Wang, Xiaofang and Juefei-Xu, Felix and Zhang, Ning and others},
  booktitle={2025 IEEE/CVF Conference on Computer Vision and Pattern Recognition (CVPR)},
  pages={18891--18901},
  year={2025},
  organization={IEEE}
}

@article{
shao2025tokencompression,
title={A Survey of Token Compression for Efficient Multimodal Large Language Models},
author={Kele Shao and Keda TAO and Kejia Zhang and Sicheng Feng and Mu Cai and Yuzhang Shang and Haoxuan You and Can Qin and Yang Sui and Huan Wang},
journal={Transactions on Machine Learning Research},
issn={2835-8856},
year={2026},
url={https://openreview.net/forum?id=G2od9JVHkE},
note={Survey Certification}
}

@inproceedings{islam2025bimba,
  title={Bimba: Selective-scan compression for long-range video question answering},
  author={Islam, Md Mohaiminul and Nagarajan, Tushar and Wang, Huiyu and Bertasius, Gedas and Torresani, Lorenzo},
  booktitle={2025 IEEE/CVF Conference on Computer Vision and Pattern Recognition (CVPR)},
  pages={29096--29107},
  year={2025},
  organization={IEEE}
}

@inproceedings{ren2025vamba,
  title={Vamba: Understanding hour-long videos with hybrid mamba-transformers},
  author={Ren, Weiming and Ma, Wentao and Yang, Huan and Wei, Cong and Zhang, Ge and Chen, Wenhu},
  booktitle={2025 IEEE/CVF International Conference on Computer Vision (ICCV)},
  pages={21197--21208},
  year={2025},
  organization={IEEE}
}

@article{liao2025mmmamba,
  title={Multimodal mamba: Decoder-only multimodal state space model via quadratic to linear distillation},
  author={Liao, Bencheng and Tao, Hongyuan and Zhang, Qian and Cheng, Tianheng and Li, Yingyue and Yin, Haoran and Liu, Wenyu and Wang, Xinggang},
  journal={arXiv preprint arXiv:2502.13145},
  year={2025}
}

@inproceedings{jiang2025storm,
  title={Storm: Token-efficient long video understanding for multimodal llms},
  author={Jiang, Jindong and Li, Xiuyu and Liu, Zhijian and Li, Muyang and Chen, Guo and Li, Zhiqi and Huang, De-An and Liu, Guilin and Yu, Zhiding and Keutzer, Kurt and others},
  booktitle={2025 IEEE/CVF International Conference on Computer Vision Workshops (ICCVW)},
  pages={5889--5900},
  year={2025},
  organization={IEEE}
}

@inproceedings{lahoti2026mamba3,
  title={Mamba-3: Improved sequence modeling using state space principles},
  author={Lahoti, Aakash Sunil and Li, Kevin and Chen, Berlin and Wang, Caitlin and Bick, Aviv and Kolter, Zico and Dao, Tri and Gu, Albert},
  booktitle={International Conference on Learning Representations},
  volume={2026},
  pages={86173--86202},
  year={2026}
}

@article{charakorn2025textlora,
  title={Text-to-lora: Instant transformer adaption},
  author={Charakorn, Rujikorn and Cetin, Edoardo and Tang, Yujin and Lange, Robert Tjarko},
  journal={arXiv preprint arXiv:2506.06105},
  year={2025}
}

@article{charakorn2026doctolora,
  title={Doc-to-lora: Learning to instantly internalize contexts},
  author={Charakorn, Rujikorn and Cetin, Edoardo and Uesaka, Shinnosuke and Lange, Robert Tjarko},
  journal={arXiv preprint arXiv:2602.15902},
  year={2026}
}

@inproceedings{zeng2025hyperedit,
  title={HyperEdit: Unlocking Instruction-based Text Editing in LLMs via Hypernetworks},
  author={Zeng, Yiming and Cao, Jinghan and Li, Zexin and Yu, Wanhao and Ye, Zhankai and Xiang, Dawei and Hua, Ting and Liu, Xin and Gao, Shangqian and Yu, Tingting},
  booktitle={Findings of the Association for Computational Linguistics: ACL 2026},
  pages={466--480},
  year={2026}
}

@inproceedings{eyuboglu2025cartridges,
  title={Cartridges: Lightweight and general-purpose long context representations via self-study},
  author={Eyuboglu, Sabri and Ehrlich, Ryan and Arora, Simran and Guha, Neel and Zinsley, Dylan and Liu, Emily and Rudra, Atri and Zou, James Y and Mirhoseini, Azalia and Re, Christopher},
  booktitle={International Conference on Learning Representations},
  volume={2026},
  pages={42642--42687},
  year={2026}
}

@article{petrov2025gistpool,
  title={Long context in-context compression by getting to the gist of gisting},
  author={Petrov, Aleksandar and Sandler, Mark and Zhmoginov, Andrey and Miller, Nolan and Vladymyrov, Max},
  journal={arXiv preprint arXiv:2504.08934},
  year={2025}
}

@article{behrouz2025titans,
  title={Titans: Learning to memorize at test time},
  author={Behrouz, Ali and Zhong, Peilin and Mirrokni, Vahab},
  journal={Advances in Neural Information Processing Systems},
  volume={38},
  pages={113506--113543},
  year={2026}
}

@inproceedings{zhang2025lact,
  title={Test-time training done right},
  author={Zhang, Tianyuan and Bi, Sai and Hong, Yicong and Zhang, Kai and Luan, Fujun and Yang, Songlin and Sunkavalli, Kalyan and Freeman, William and Tan, Hao},
  booktitle={International Conference on Learning Representations},
  volume={2026},
  pages={157604--157638},
  year={2026}
}

@inproceedings{shin2025gpi,
  title={Generative prompt internalization},
  author={Shin, Haebin and Ji, Lei and Gong, Yeyun and Kim, Sungdong and Choi, Eunbi and Seo, Minjoon},
  booktitle={Proceedings of the 2025 Conference of the Nations of the Americas Chapter of the Association for Computational Linguistics: Human Language Technologies (Volume 1: Long Papers)},
  pages={7338--7363},
  year={2025}
}

@article{yang2025keyevl,
  title={Kwai keye-vl 1.5 technical report},
  author={Yang, Biao and Wen, Bin and Ding, Boyang and Liu, Changyi and Chu, Chenglong and Song, Chengru and Rao, Chongling and Yi, Chuan and Li, Da and Zang, Dunju and others},
  journal={arXiv preprint arXiv:2509.01563},
  year={2025}
}

@inproceedings{yu2025minicpmv45,
  title={Minicpm-v 4.5: Cooking efficient mllms via architecture, data, and training recipe},
  author={Yu, Tianyu and Wang, Zefan and Wang, Chongyi and Huang, Fuwei and Ma, Wenshuo and He, Zhihui and Cai, Tianchi and Chen, Weize and Huang, Yuxiang and Zhao, Ranchi and others},
  booktitle={Proceedings of the IEEE/CVF Conference on Computer Vision and Pattern Recognition},
  pages={11704--11715},
  year={2026}
}

@inproceedings{li2024videochatflash,
  title={Videochat-flash: Hierarchical compression for long-context video modeling},
  author={Li, Xinhao and Wang, Yi and Yu, Jiashuo and Zeng, Xiangyu and Zhu, Yuhan and Huang, Haian and Gao, Jianfei and Li, Kunchang and He, Yinan and Wang, Chenting and others},
  booktitle={International Conference on Learning Representations},
  volume={2026},
  pages={109089--109117},
  year={2026}
}

@article{wang2025internvl35,
  title={Internvl3. 5: Advancing open-source multimodal models in versatility, reasoning, and efficiency},
  author={Wang, Weiyun and Gao, Zhangwei and Gu, Lixin and Pu, Hengjun and Cui, Long and Wei, Xingguang and Liu, Zhaoyang and Jing, Linglin and Ye, Shenglong and Shao, Jie and others},
  journal={arXiv preprint arXiv:2508.18265},
  year={2025}
}

@article{cho2025perceptionlm,
  title={Perceptionlm: Open-access data and models for detailed visual understanding},
  author={Cho, Jang Hyun and Madotto, Andrea and Mavroudi, Effrosyni and Afouras, Triantafyllos and Nagarajan, Tushar and Maaz, Muhammad and Song, Yale and Ma, Tengyu and Hu, Shuming and Jain, Suyog and others},
  journal={Advances in Neural Information Processing Systems},
  volume={38},
  pages={23475--23537},
  year={2026}
}

@misc{hong2025glm41v,
      title={GLM-4.5V and GLM-4.1V-Thinking: Towards Versatile Multimodal Reasoning with Scalable Reinforcement Learning}, 
      author={V Team and Wenyi Hong and Wenmeng Yu and Xiaotao Gu and Guo Wang and Guobing Gan and Haomiao Tang and Jiale Cheng and Ji Qi and Junhui Ji and Lihang Pan and Shuaiqi Duan and Weihan Wang and Yan Wang and Yean Cheng and Zehai He and Zhe Su and Zhen Yang and Ziyang Pan and Aohan Zeng and Baoxu Wang and Bin Chen and Boyan Shi and Changyu Pang and Chenhui Zhang and Da Yin and Fan Yang and Guoqing Chen and Haochen Li and Jiale Zhu and Jiali Chen and Jiaxing Xu and Jiazheng Xu and Jing Chen and Jinghao Lin and Jinhao Chen and Jinjiang Wang and Junjie Chen and Leqi Lei and Letian Gong and Leyi Pan and Mingdao Liu and Mingde Xu and Mingzhi Zhang and Qinkai Zheng and Ruiliang Lyu and Shangqin Tu and Sheng Yang and Shengbiao Meng and Shi Zhong and Shiyu Huang and Shuyuan Zhao and Siyan Xue and Tianshu Zhang and Tianwei Luo and Tianxiang Hao and Tianyu Tong and Wei Jia and Wenkai Li and Xiao Liu and Xiaohan Zhang and Xin Lyu and Xinyu Zhang and Xinyue Fan and Xuancheng Huang and Yadong Xue and Yanfeng Wang and Yanling Wang and Yanzi Wang and Yifan An and Yifan Du and Yiheng Huang and Yilin Niu and Yiming Shi and Yu Wang and Yuan Wang and Yuanchang Yue and Yuchen Li and Yusen Liu and Yutao Zhang and Yuting Wang and Yuxuan Zhang and Zhao Xue and Zhengxiao Du and Zhenyu Hou and Zihan Wang and Peng Zhang and Debing Liu and Bin Xu and Juanzi Li and Minlie Huang and Yuxiao Dong and Jie Tang},
      year={2026},
      eprint={2507.01006},
      archivePrefix={arXiv},
      primaryClass={cs.CV},
      url={https://arxiv.org/abs/2507.01006}, 
}

@misc{chen2025eagle25,
      title={Eagle 2.5: Boosting Long-Context Post-Training for Frontier Vision-Language Models}, 
      author={Guo Chen and Zhiqi Li and Shihao Wang and Jindong Jiang and Yicheng Liu and Lidong Lu and De-An Huang and Wonmin Byeon and Matthieu Le and Tuomas Rintamaki and Tyler Poon and Max Ehrlich and Tuomas Rintamaki and Tyler Poon and Tong Lu and Limin Wang and Bryan Catanzaro and Jan Kautz and Andrew Tao and Zhiding Yu and Guilin Liu},
      year={2025},
      eprint={2504.15271},
      archivePrefix={arXiv},
      primaryClass={cs.CV},
      url={https://arxiv.org/abs/2504.15271}, 
}

@inproceedings{
patraucean2023perceptiontest,
title={Perception Test: A Diagnostic Benchmark for Multimodal Video Models},
author={Viorica Patraucean and Lucas Smaira and Ankush Gupta and Adria Recasens Continente and Larisa Markeeva and Dylan Sunil Banarse and Skanda Koppula and Joseph Heyward and Mateusz Malinowski and Yi Yang and Carl Doersch and Tatiana Matejovicova and Yury Sulsky and Antoine Miech and Alexandre Fr{\'e}chette and Hanna Klimczak and Raphael Koster and Junlin Zhang and Stephanie Winkler and Yusuf Aytar and Simon Osindero and Dima Damen and Andrew Zisserman and Joao Carreira},
booktitle={Thirty-seventh Conference on Neural Information Processing Systems Datasets and Benchmarks Track},
year={2023},
url={https://openreview.net/forum?id=HYEGXFnPoq}
}

@InProceedings{xiao2021nextqa,
    author    = {Xiao, Junbin and Shang, Xindi and Yao, Angela and Chua, Tat-Seng},
    title     = {NExT-QA: Next Phase of Question-Answering to Explaining Temporal Actions},
    booktitle = {Proceedings of the IEEE/CVF Conference on Computer Vision and Pattern Recognition (CVPR)},
    month     = {June},
    year      = {2021},
    pages     = {9777-9786}
}

@misc{wang2025lvbench,
      title={LVBench: An Extreme Long Video Understanding Benchmark}, 
      author={Weihan Wang and Zehai He and Wenyi Hong and Yean Cheng and Xiaohan Zhang and Ji Qi and Shiyu Huang and Bin Xu and Yuxiao Dong and Ming Ding and Jie Tang},
      year={2024},
      eprint={2406.08035},
      archivePrefix={arXiv},
      primaryClass={cs.CV}
}

@misc{li2024mvbench,
      title={MVBench: A Comprehensive Multi-modal Video Understanding Benchmark}, 
      author={Kunchang Li and Yali Wang and Yinan He and Yizhuo Li and Yi Wang and Yi Liu and Zun Wang and Jilan Xu and Guo Chen and Ping Luo and Limin Wang and Yu Qiao},
      year={2024},
      eprint={2311.17005},
      archivePrefix={arXiv},
      primaryClass={cs.CV},
      url={https://arxiv.org/abs/2311.17005}, 
}

@misc{shangguan2025tomato,
      title={TOMATO: Assessing Visual Temporal Reasoning Capabilities in Multimodal Foundation Models}, 
      author={Ziyao Shangguan and Chuhan Li and Yuxuan Ding and Yanan Zheng and Yilun Zhao and Tesca Fitzgerald and Arman Cohan},
      year={2024},
      eprint={2410.23266},
      archivePrefix={arXiv},
      primaryClass={cs.CV},
      url={https://arxiv.org/abs/2410.23266}, 
}

@article{comanici2025gemini25,
  title={Gemini 2.5: Pushing the frontier with advanced reasoning, multimodality, long context, and next generation agentic capabilities},
  author={Comanici, Gheorghe and Bieber, Eric and Schaekermann, Mike and Pasupat, Ice and Sachdeva, Noveen and Dhillon, Inderjit and Blistein, Marcel and Ram, Ori and Zhang, Dan and Rosen, Evan and others},
  journal={arXiv preprint arXiv:2507.06261},
  year={2025}
}

@misc{google2025gemini3,
  title={{Gemini 3 Pro} Model Card},
  author={Google},
  url={https://storage.googleapis.com/deepmind-media/Model-Cards/Gemini-3-Pro-Model-Card.pdf},
  year={2025}
}

@misc{openai2025gpt5,
  title={{GPT-5} System Card},
  author={OpenAI},
  url={https://openai.com/index/gpt-5-system-card/},
  year={2025}
}

@misc{anthropic2025sonnet45,
  title={Claude Sonnet 4.5 System Card},
  author={Anthropic},
  year={2025},
  url={https://assets.anthropic.com/m/12f214efcc2f457a/original/Claude-Sonnet-4-5-System-Card.pdf}
}

@misc{molmo2team2026,
      title={Molmo2: Open Weights and Data for Vision-Language Models with Video Understanding and Grounding}, 
      author={Christopher Clark and Jieyu Zhang and Zixian Ma and Jae Sung Park and Mohammadreza Salehi and Rohun Tripathi and Sangho Lee and Zhongzheng Ren and Chris Dongjoo Kim and Yinuo Yang and Vincent Shao and Yue Yang and Weikai Huang and Ziqi Gao and Taira Anderson and Jianrui Zhang and Jitesh Jain and George Stoica and Winson Han and Ali Farhadi and Ranjay Krishna},
      year={2026},
      eprint={2601.10611},
      archivePrefix={arXiv},
      primaryClass={cs.CV},
      url={https://arxiv.org/abs/2601.10611}, 
}

@misc{liu2026infomerge,
      title={InfoMerge: Information-aware Token Compression for Efficient Video Large Language Models}, 
      author={Xinxin Liu and Shiwei Gan and Xiao Liu and Yafeng Yin and Lei Xie and Sanglu Lu},
      year={2026},
      eprint={2606.02161},
      archivePrefix={arXiv},
      primaryClass={cs.CV},
      url={https://arxiv.org/abs/2606.02161}, 
}

@inproceedings{
zhang2026xcomp,
title={One Token per Highly Selective Frame: Towards Extreme Compression for Long Video Understanding},
author={Zheyu Aqa Zhang and Ziqi Pang and Shixing Chen and Xiang Hao and Vimal Bhat and Yu-Xiong Wang},
booktitle={The Thirty-ninth Annual Conference on Neural Information Processing Systems},
year={2025},
}

@inproceedings{
kim2025infinipotv,
title={InfiniPot-V: Memory-Constrained {KV} Cache Compression for Streaming Video Understanding},
author={Minsoo Kim and Kyuhong Shim and Jungwook Choi and Simyung Chang},
booktitle={The Thirty-ninth Annual Conference on Neural Information Processing Systems},
year={2025},
url={https://openreview.net/forum?id=hFxOZjHyTg}
}

@inproceedings{wang2025adaretake,
    title = "{A}da{R}e{T}a{K}e: Adaptive Redundancy Reduction to Perceive Longer for Video-language Understanding",
    author = "Wang, Xiao  and
      Si, Qingyi  and
      Zhu, Shiyu  and
      Wu, Jianlong  and
      Cao, Li  and
      Nie, Liqiang",
    booktitle = "Findings of the Association for Computational Linguistics: ACL 2025",
    month = jul,
    year = "2025",
    address = "Vienna, Austria",
    publisher = "Association for Computational Linguistics",
    pages = "5417--5432",
}

@inproceedings{
liu2026shine,
title={{SHINE}: A Scalable In-Context Hypernetwork for Mapping Context to Lo{RA} in a Single Pass},
author={Yewei Liu and Xiyuan Wang and Yansheng Mao and Yoav Gelberg and Haggai Maron and Muhan Zhang},
booktitle={Forty-third International Conference on Machine Learning},
year={2026},
url={https://openreview.net/forum?id=ZMexYcAibv}
}

@misc{
abdalla2025zhyper,
title={Zhyper: Factorized Hypernetworks for Conditioned {LLM} Fine-Tuning},
author={M. H. I. Abdalla and Zhipin Wang and Christian Frey and Steffen Eger and Josif Grabocka},
year={2026},
url={https://openreview.net/forum?id=p0hiJ09B2V}
}

@misc{zeng2026moegen,
      title={MoEGen: Mixture-of-Experts for Instance-Adaptive LoRA Generation}, 
      author={Yiming Zeng and Lei Lu and Zexin Li and Zhuochun Li and Shuoqiu Li and Shuyi Liao and Xidong Wu and Zeyu Zhang and Minmei Wang and Yu Zhao and Tingting Yu and Shangqian Gao},
      year={2026},
      eprint={2608.03275},
      archivePrefix={arXiv},
      primaryClass={cs.CL},
      url={https://arxiv.org/abs/2608.03275}, 
}

@misc{han2026weightsurvey,
      title={A Survey of Weight Space Learning: Understanding, Representation, and Generation}, 
      author={Xiaolong Han and Zehong Wang and Bo Zhao and Binchi Zhang and Jundong Li and Damian Borth and Rose Yu and Haggai Maron and Yanfang Ye and Lu Yin and Ferrante Neri},
      year={2026},
      eprint={2603.10090},
      archivePrefix={arXiv},
      primaryClass={cs.LG},
      url={https://arxiv.org/abs/2603.10090}, 
}

@misc{
zeng2025genweights,
title={Generative Modeling of Weights: Generalization or Memorization?},
author={Boya Zeng and Yida Yin and Zhiqiu Xu and Zhuang Liu},
year={2025},
url={https://openreview.net/forum?id=QbSeApdgfi}
}

@misc{wang2026learnwhat,
      title={Learning What to Remember: Test-Time Training via Context Distillation}, 
      author={Zixuan Wang and Xingyu Dang and Rui-Jie Zhu and Zixin Wen and Hengyu Fu and Wenhao Chai and Jason D. Lee},
      year={2026},
      eprint={2608.01672},
      archivePrefix={arXiv},
      primaryClass={cs.CL},
      url={https://arxiv.org/abs/2608.01672}, 
}

@inproceedings{
cheng2025vilamp,
title={Scaling Video-Language Models to 10K Frames via Hierarchical Differential Distillation},
author={Chuanqi Cheng and Jian Guan and Wei Wu and Rui Yan},
booktitle={Forty-second International Conference on Machine Learning},
year={2025},
url={https://openreview.net/forum?id=1CK1kuH1he}
}

@inproceedings{
song2025videonsa,
title={Video{NSA}: Native Sparse Attention Scales Video Understanding},
author={Enxin Song and Wenhao Chai and Shusheng Yang and Ethan J. Armand and Xiaojun Shan and Haiyang Xu and Jianwen Xie and Zhuowen Tu},
booktitle={The Fourteenth International Conference on Learning Representations},
year={2026},
url={https://openreview.net/forum?id=zA2LbsUMDd}
}

@inproceedings{tan2025instant,
    title = "Instant Personalized Large Language Model Adaptation via Hypernetwork",
    author = "Tan, Zhaoxuan  and
      Zhang, Zixuan  and
      Wen, Haoyang  and
      Li, Zheng  and
      Zhang, Rongzhi  and
      Chen, Pei  and
      Mo, Fengran  and
      Liu, Zheyuan  and
      Zeng, Qingkai  and
      Yin, Qingyu  and
      Jiang, Meng",

    booktitle = "Proceedings of the 64th Annual Meeting of the {A}ssociation for {C}omputational {L}inguistics (Volume 1: Long Papers)",
    month = jul,
    year = "2026",
    address = "San Diego, California, United States",
    publisher = "Association for Computational Linguistics",
    pages = "23557--23580",
}

@inproceedings{
ding2026videozoomer,
title={VideoZoomer: Reinforcement-Learned Temporal Focusing for Long Video Reasoning},
author={Yang Ding and Xin Lai and Yizhen Zhang and Wei Li and Ruihang Chu and Yujiu Yang},
booktitle={The Fourteenth International Conference on Learning Representations},
year={2026},
url={https://openreview.net/forum?id=ARHCFvgx6G}
}

@misc{kim2026swift,
      title={Swift Sampling: Selecting Temporal Surprises via Taylor Series}, 
      author={Dahye Kim and Bhuvan Sachdeva and Karan Uppal and Naman Gupta and Vineeth N. Balasubramanian and Deepti Ghadiyaram},
      year={2026},
      eprint={2605.22678},
      archivePrefix={arXiv},
      primaryClass={cs.CV},
      url={https://arxiv.org/abs/2605.22678}, 
}

@misc{zhang2025llavamini,
      title={LLaVA-Mini: Efficient Image and Video Large Multimodal Models with One Vision Token}, 
      author={Shaolei Zhang and Qingkai Fang and Zhe Yang and Yang Feng},
      year={2025},
      eprint={2501.03895},
      archivePrefix={arXiv},
      primaryClass={cs.CV},
      url={https://arxiv.org/abs/2501.03895}, 
}

@misc{liu2024nvila,
      title={NVILA: Efficient Frontier Visual Language Models}, 
      author={Zhijian Liu and Ligeng Zhu and Baifeng Shi and Zhuoyang Zhang and Yuming Lou and Shang Yang and Haocheng Xi and Shiyi Cao and Yuxian Gu and Dacheng Li and Xiuyu Li and Yunhao Fang and Yukang Chen and Cheng-Yu Hsieh and De-An Huang and An-Chieh Cheng and Vishwesh Nath and Jinyi Hu and Sifei Liu and Ranjay Krishna and Daguang Xu and Xiaolong Wang and Pavlo Molchanov and Jan Kautz and Hongxu Yin and Song Han and Yao Lu},
      year={2024},
      eprint={2412.04468},
      archivePrefix={arXiv},
      primaryClass={cs.CV},
      url={https://arxiv.org/abs/2412.04468}, 
}

@article{
li2024llavaonevision,
title={{LL}a{VA}-OneVision: Easy Visual Task Transfer},
author={Bo Li and Yuanhan Zhang and Dong Guo and Renrui Zhang and Feng Li and Hao Zhang and Kaichen Zhang and Peiyuan Zhang and Yanwei Li and Ziwei Liu and Chunyuan Li},
journal={Transactions on Machine Learning Research},
issn={2835-8856},
year={2025},
url={https://openreview.net/forum?id=zKv8qULV6n},
note={}
}

@misc{wang2025internvideo25,
      title={InternVideo2.5: Empowering Video MLLMs with Long and Rich Context Modeling}, 
      author={Yi Wang and Xinhao Li and Ziang Yan and Yinan He and Jiashuo Yu and Xiangyu Zeng and Chenting Wang and Changlian Ma and Haian Huang and Jianfei Gao and Min Dou and Kai Chen and Wenhai Wang and Yu Qiao and Yali Wang and Limin Wang},
      year={2025},
      eprint={2501.12386},
      archivePrefix={arXiv},
      primaryClass={cs.CV},
      url={https://arxiv.org/abs/2501.12386}, 
}

@article{cheng2024videollama2,
  title={VideoLLaMA 2: Advancing Spatial-Temporal Modeling and Audio Understanding in Video-LLMs},
  author={Cheng, Zesen and Leng, Sicong and Zhang, Hang and Xin, Yifei and Li, Xin and Chen, Guanzheng and Zhu, Yongxin and Zhang, Wenqi and Luo, Ziyang and Zhao, Deli and Bing, Lidong},
  journal={arXiv preprint arXiv:2406.07476},
  year={2024},
  url = {https://arxiv.org/abs/2406.07476}
}

@inproceedings{
ye2024mplugowl3,
title={m{PLUG}-Owl3: Towards Long Image-Sequence Understanding in Multi-Modal Large Language Models},
author={Jiabo Ye and Haiyang Xu and Haowei Liu and Anwen Hu and Ming Yan and Qi Qian and Ji Zhang and Fei Huang and Jingren Zhou},
booktitle={The Thirteenth International Conference on Learning Representations},
year={2025},
url={https://openreview.net/forum?id=pr37sbuhVa}
}

@article{liu2024kangaroo,
	title={Kangaroo: A Powerful Video-Language Model Supporting Long-context Video Input},
	author={Liu, Jiajun and Wang, Yibing and Ma, Hanghang and Wu, Xiaoping and Ma, Xiaoqi and Wei, xiaoming and Jiao, Jianbin and Wu, Enhua and Hu, Jie},
	journal={arXiv preprint arXiv:2408.15542},
	year={2024}
}

@misc{zhang2025videollama3,
      title={VideoLLaMA 3: Frontier Multimodal Foundation Models for Image and Video Understanding}, 
      author={Boqiang Zhang and Kehan Li and Zesen Cheng and Zhiqiang Hu and Yuqian Yuan and Guanzheng Chen and Sicong Leng and Yuming Jiang and Hang Zhang and Xin Li and Peng Jin and Wenqi Zhang and Fan Wang and Lidong Bing and Deli Zhao},
      year={2025},
      eprint={2501.13106},
      archivePrefix={arXiv},
      primaryClass={cs.CV},
      url={https://arxiv.org/abs/2501.13106}, 
}

@inproceedings{
alayrac2022flamingo,
title={Flamingo: a Visual Language Model for Few-Shot Learning},
author={Jean-Baptiste Alayrac and Jeff Donahue and Pauline Luc and Antoine Miech and Iain Barr and Yana Hasson and Karel Lenc and Arthur Mensch and Katherine Millican and Malcolm Reynolds and Roman Ring and Eliza Rutherford and Serkan Cabi and Tengda Han and Zhitao Gong and Sina Samangooei and Marianne Monteiro and Jacob Menick and Sebastian Borgeaud and Andrew Brock and Aida Nematzadeh and Sahand Sharifzadeh and Mikolaj Binkowski and Ricardo Barreira and Oriol Vinyals and Andrew Zisserman and Karen Simonyan},
booktitle={Advances in Neural Information Processing Systems},
editor={Alice H. Oh and Alekh Agarwal and Danielle Belgrave and Kyunghyun Cho},
year={2022},
url={https://openreview.net/forum?id=EbMuimAbPbs}
}

@inproceedings{li2023blip2,
author = {Li, Junnan and Li, Dongxu and Savarese, Silvio and Hoi, Steven},
title = {BLIP-2: bootstrapping language-image pre-training with frozen image encoders and large language models},
year = {2023},
publisher = {JMLR.org},
booktitle = {Proceedings of the 40th International Conference on Machine Learning},
articleno = {814},
numpages = {13},
location = {Honolulu, Hawaii, USA},
series = {ICML'23}
}

@misc{jaegle2021perceiver,
      title={Perceiver: General Perception with Iterative Attention}, 
      author={Andrew Jaegle and Felix Gimeno and Andrew Brock and Andrew Zisserman and Oriol Vinyals and Joao Carreira},
      year={2021},
      eprint={2103.03206},
      archivePrefix={arXiv},
      primaryClass={cs.CV},
      url={https://arxiv.org/abs/2103.03206}, 
}

@inproceedings{
bolya2023tome,
title={Token Merging: Your ViT But Faster},
author={Daniel Bolya and Cheng-Yang Fu and Xiaoliang Dai and Peizhao Zhang and Christoph Feichtenhofer and Judy Hoffman},
booktitle={The Eleventh International Conference on Learning Representations },
year={2023},
url={https://openreview.net/forum?id=JroZRaRw7Eu}
}

@article{xiao2024streamingllm,
        title={Efficient Streaming Language Models with Attention Sinks},
        author={Xiao, Guangxuan and Tian, Yuandong and Chen, Beidi and Han, Song and Lewis, Mike},
        journal={arXiv},
        year={2023}
        }

@inproceedings{zhang2023h2o,
author = {Zhang, Zhenyu and Sheng, Ying and Zhou, Tianyi and Chen, Tianlong and Zheng, Lianmin and Cai, Ruisi and Song, Zhao and Tian, Yuandong and R{\'e}, Christopher and Barrett, Clark and Wang, Zhangyang and Chen, Beidi},
title = {H2O: heavy-hitter oracle for efficient generative inference of large language models},
year = {2023},
publisher = {Curran Associates Inc.},
address = {Red Hook, NY, USA},
booktitle = {Proceedings of the 37th International Conference on Neural Information Processing Systems},
articleno = {1506},
numpages = {50},
location = {New Orleans, LA, USA},
series = {NIPS '23}
}

@inproceedings{li2021prefix,
    title = "Prefix-Tuning: Optimizing Continuous Prompts for Generation",
    author = "Li, Xiang Lisa  and
      Liang, Percy",
    booktitle = "Proceedings of the 59th Annual Meeting of the Association for Computational Linguistics and the 11th International Joint Conference on Natural Language Processing (Volume 1: Long Papers)",
    month = aug,
    year = "2021",
    address = "Online",
    publisher = "Association for Computational Linguistics",
    url = "https://aclanthology.org/2021.acl-long.353/",
    doi = "10.18653/v1/2021.acl-long.353",
    pages = "4582--4597",
}

@inproceedings{mahabadi2021hyperformer,
    title = "Parameter-efficient Multi-task Fine-tuning for Transformers via Shared Hypernetworks",
    author = "Karimi Mahabadi, Rabeeh  and
      Ruder, Sebastian  and
      Dehghani, Mostafa  and
      Henderson, James",
    booktitle = "Proceedings of the 59th Annual Meeting of the Association for Computational Linguistics and the 11th International Joint Conference on Natural Language Processing (Volume 1: Long Papers)",
    month = aug,
    year = "2021",
    address = "Online",
    publisher = "Association for Computational Linguistics",
    url = "https://aclanthology.org/2021.acl-long.47/",
    doi = "10.18653/v1/2021.acl-long.47",
    pages = "565--576",
}

@inproceedings{ba2016fastweights,
author = {Ba, Jimmy and Hinton, Geoffrey and Mnih, Volodymyr and Leibo, Joel Z. and Ionescu, Catalin},
title = {Using fast weights to attend to the recent past},
year = {2016},
isbn = {9781510838819},
publisher = {Curran Associates Inc.},
address = {Red Hook, NY, USA},
booktitle = {Proceedings of the 30th International Conference on Neural Information Processing Systems},
pages = {4338–4346},
numpages = {9},
location = {Barcelona, Spain},
series = {NIPS'16}
}

@article{schmidhuber1992fastweights,
author = {Schmidhuber, J{\"u}rgen},
title = {Learning to control fast-weight memories: an alternative to dynamic recurrent networks},
year = {1992},
issue_date = {Jan. 1992},
publisher = {MIT Press},
address = {Cambridge, MA, USA},
volume = {4},
number = {1},
issn = {0899-7667},
url = {https://doi.org/10.1162/neco.1992.4.1.131},
doi = {10.1162/neco.1992.4.1.131},
journal = {Neural Comput.},
month = jan,
pages = {131–139},
numpages = {9}
}

@misc{gu2022s4,
      title={Efficiently Modeling Long Sequences with Structured State Spaces}, 
      author={Albert Gu and Karan Goel and Christopher Ré},
      year={2022},
      eprint={2111.00396},
      archivePrefix={arXiv},
      primaryClass={cs.LG},
      url={https://arxiv.org/abs/2111.00396}, 
}

@inproceedings{
zhang2025sparsevlm,
title={Sparse{VLM}: Visual Token Sparsification for Efficient Vision-Language Model Inference},
author={Yuan Zhang and Chun-Kai Fan and Junpeng Ma and Wenzhao Zheng and Tao Huang and Kuan Cheng and Denis A Gudovskiy and Tomoyuki Okuno and Yohei Nakata and Kurt Keutzer and Shanghang Zhang},
booktitle={Forty-second International Conference on Machine Learning},
year={2025},
url={https://openreview.net/forum?id=80faIPZ67S}
}

@misc{xing2025pyramiddrop,
      title={PyramidDrop: Accelerating Your Large Vision-Language Models via Pyramid Visual Redundancy Reduction}, 
      author={Long Xing and Qidong Huang and Xiaoyi Dong and Jiajie Lu and Pan Zhang and Yuhang Zang and Yuhang Cao and Conghui He and Jiaqi Wang and Feng Wu and Dahua Lin},
      year={2025},
      eprint={2410.17247},
      archivePrefix={arXiv},
      primaryClass={cs.CV},
      url={https://arxiv.org/abs/2410.17247}, 
}

\appendix
\clearpage
\section{Baseline implementations and comparison scope}
\label{app:baselines}

\noindent\textbf{Comparison scope.} We group the efficient models into a
matched budget set that we reevaluate at 256 uniformly sampled frames under one
protocol (BIMBA, Vamba, VideoNSA, and \method{}) and a reported set whose scores are
copied from each method's own paper. The matched budget set fixes the evaluation
budget (frame count, decoding, hardware, and suite), so these rows isolate an
evaluation budget rather than the compression mechanism. Architecture only
conclusions are therefore not warranted across every row, because checkpoints,
vision and language backbones, training mixtures, sampling rules, spatial
resolution, visual token budgets, prompting, and decoding can differ. For the
reported set we take only the MLVU and LVBench \emph{test} splits and leave a cell
blank when a paper reports only the MLVU dev or M-avg number, and we compute
aggregate scores only for rows that cover the complete suite. Several concurrent
open video language models (InternVideo2.5~\citep{wang2025internvideo25}, MiniCPM-V
4.5~\citep{yu2025minicpmv45}, and InternVL3.5~\citep{wang2025internvl35}) report on
overlapping suites but under their own frame budgets, backbones, and training
mixtures, so we describe rather than tabulate them as budget matched rows. The
following descriptions identify the mechanism represented by each baseline and its
principal comparison qualifier.

\subsection{Open reference models}

\paragraph{Qwen3-VL-8B.} Qwen3-VL combines a SigLIP-2 vision encoder, a two layer
vision language merger, and a Qwen3 decoder~\citep{bai2025qwen3vl}. Interleaved
MRoPE, DeepStack visual injection, and timestamp text support multimodal temporal
representation. It is an open full context reference rather than a long video
compression method. The 8B release includes Instruct and Thinking variants, so a
comparison must specify the checkpoint, reasoning mode, frame sampling, visual
budget, and prompting protocol.

\paragraph{Eagle2.5-8B.} Eagle 2.5 projects SigLIP visual features into a Qwen2.5
language model through an MLP and uses LLaVA style image
tiling~\citep{chen2025eagle25}. Its Image Area Preservation and Automatic Degradation
Sampling preserve aspect ratio and allocate the input length budget between text and
visual tokens. It is a long context reference rather than a temporal compressor.
Reported Video-MME settings include 512 frames, so backbone, frame count, and
resolution differ from a fixed 256 frame comparison.

\paragraph{PerceptionLM (PLM-8B).} PerceptionLM links a pretrained Perception
Encoder to a Llama 3 decoder using a two layer MLP
projector~\citep{cho2025perceptionlm}. Its 8B variant uses a PE G/14 encoder and Llama
3.1 8B. For video, it retains sampled frames and applies spatial pooling within each
frame rather than selecting a shorter temporal sequence. Its visual backbone,
resolution, and frame protocol differ from the efficient model configurations here.

\paragraph{LLaVA-Video-7B.} LLaVA-Video extends LLaVA-OneVision with synthetic
video instruction data and uses SigLIP with a Qwen2 language
model~\citep{zhang2024llavavideo}. Its optional SlowFast representation allocates
different spatial pooling rates to slow and fast frame groups, allowing broader
temporal coverage at a bounded token count. Several published benchmark results
are marked for overlap between a corresponding dataset and the training mixture, so
such scores are not strictly independent comparisons.

\paragraph{LLaVA-OneVision-7B.} LLaVA-OneVision is an open full context model on a
Qwen2-7B backbone that unifies single image, multi image, and video
transfer~\citep{li2024llavaonevision}. Each frame is token reduced by bilinear
interpolation and evaluated at 32 frames. It is an open reference rather than a
token compression method, and it does not report TOMATO, MLVU test, or LVBench.

\paragraph{Molmo2-8B.} Molmo2 encodes independently sampled video frames with a
SigLIP-2 style vision transformer and connector, spatially pools video patches, and
interleaves the resulting visual tokens with timestamp text~\citep{molmo2team2026}.
It uses pretraining, multitask supervised finetuning, and a long context stage.
The reported video protocol uses fixed sampling and frame caps, which materially
affect context length and compute. It is consequently an open reference, not an
architecture matched token compression baseline.

\paragraph{VideoChat2.} VideoChat2 couples a UMT-L visual encoder and a QFormer with a
7B language backbone and is the reference model introduced with the MVBench
benchmark~\citep{li2024mvbench}. It is widely retabulated in later long video
evaluations. We report its Mistral-7B numbers under uniform 16 frame sampling and use
its zero shot scores, so its MVBench value is comparatively strong while its
long video Video-MME scores are modest.

\subsection{Efficient video language models}

\paragraph{BIMBA.} BIMBA jointly processes frame derived visual features and a
smaller set of visual queries, interleaves those queries through the sequence, and
uses a bidirectional selective scan before sending compressed tokens to an
LLM~\citep{islam2025bimba}. This reduces the LLM facing context while replacing
full sequence attention in the compressor with linear time sequence processing. Its
paper includes a shared data, shared decoder comparison, but cross paper rows can
otherwise differ in both decoder and training data.

\paragraph{Vamba.} Vamba avoids visual token reduction. It retains transformer
self attention for text, replaces video token causal self attention with Mamba2,
and connects modalities through text query/video key value
cross attention~\citep{ren2025vamba}. Its efficiency comes from linear recurrent
processing over video tokens rather than a shorter visual prefix. The reported runtime
comparison uses a shared Qwen2-VL setup, but Vamba adds substantial parameters and uses
different training and inference frame capacities.

\paragraph{VideoNSA.} VideoNSA applies learnable Native Sparse Attention to vision
tokens in Qwen2.5-VL-7B~\citep{bai2025qwen25vl} while retaining standard grouped query
attention for text~\citep{song2025videonsa}. Gated branches compress blocks, select
salient blocks, and retain a sliding local window, reducing vision key value
interactions without all to all visual attention. Its main model is end to end
finetuned on a filtered LLaVA-Video subset, a training condition that should be kept
separate from methods evaluated with their released configurations.

\section{Implementation details}
\label{app:implementation}

\paragraph{Selective scan recurrence.}
The state space layer maintains a state $\mathbf h_i\in\mathbb{R}^{N\times d'}$,
where $N=64$ is the SSM state size and $d'$ is the per channel model width, taken
equal to the hidden width $d$ up to the Mamba2 expansion factor. Writing $z_i$ for
the $i$-th row of the scan input $\mathbf Z$ (distinct from the hypernetwork latent
$\mathbf z^{(\ell)}$ of Eq.~\eqref{eq:conditioning}), the state updates with
input dependent gates,
\begin{equation}
 \mathbf h_i=\bar A(z_i)\odot\mathbf h_{i-1}+\bar B(z_i)\,z_i^{\top},\qquad
 y_i=C(z_i)^\top\mathbf h_i+D\odot z_i,
\end{equation}
where $\bar B(z_i)\in\mathbb R^{N}$, so that $\bar B(z_i)\,z_i^{\top}$ is the
rank one injection of the input into the state. We use the state space duality
Mamba2 formulation with state size $N=64$ and separate parameters for the forward
and reverse scans. The scan output projection is initialized from
$\mathcal N(0,(3\times10^{-4})^2)$, which preserves the bilinear resampling
initialization while allowing gradients to reach the scan interior from the first
step.

\paragraph{Adapter head shapes.}
Mirroring Eq.~\eqref{eq:lora-factors}, the module specific projection
$\mathbf P_m^{(\ell)}\in\mathbb R^{rd_z\times d_z}$ reshapes the module latent into
$\mathbf Z_m^{(\ell)}\in\mathbb R^{r\times d_z}$, and the mixing matrices
$\mathbf U_m^{(\ell)}\in\mathbb R^{d_z\times d_{\mathrm{in},m}}$ and
$\mathbf V_m^{(\ell)}\in\mathbb R^{d_z\times d_{\mathrm{out},m}}$ map it to the LoRA
factors $\mathbf A_m^{(\ell)}$ and $\mathbf B_m^{(\ell)}$. The
$\operatorname{RowNorm}$ operator normalizes $\mathbf Z_m^{(\ell)}$ row by row
before the mixing step, keeping the generated rank one directions at a stable scale.

\paragraph{DeepStack alignment.}
Qwen3-VL injects visual encoder features at several early decoder layers. We pool
these features using the same segment map as Eq.~\eqref{eq:keepset}, keeping every
DeepStack feature aligned with the retained token that represents its segment.

\paragraph{Efficient two pass training.}
The vocabulary has 151,936 entries, so full sequence teacher and student logits
are not materialized. Instead, both passes return final hidden states and the
frozen output head forms logits only at response positions, in chunks. Teacher and
student use identical pixels, and because the vision encoder is frozen, its features
are computed once and reused by the student in detached form. Under activation
checkpointing, the student forward must remain the final forward before backward
so recomputation retrieves the correct contraction state.

\paragraph{Paired update analysis protocol.}
For every evaluation question in Table~\ref{tab:weight-evidence}, the compressor
constructs one contracted decoder input from the same sampled video frames and
prompt. This input includes the same carrier tokens in both conditions. We then
run the same frozen decoder twice: once with the LoRA update produced by the
hypernetworks and once with that update discarded. The response parser and decoding
rule are fixed. The paired accuracy difference therefore measures what the generated
update adds on top of the same compressed visual context.

\paragraph{Latency measurements.}
Table~\ref{tab:latency-grid} reports prefill, time to first token, and per token
decode latency for the base backbone and \method{} at $k\in\{6,16\}$ across
128 to 1024 frames. Prefill falls by $3.35\times$ at 256 frames and $4.14\times$ at
1024 frames, while per token decode latency is essentially unchanged because the
generated update is merged into the frozen weights before decoding.

\clearpage
\begin{table*}[!t]
\caption{\textbf{Video question answering accuracy, full comparison}, in
  percent. Table~\ref{tab:main} in the main text is the subset of these rows
  most directly comparable to \method{}. As in
  Table~\ref{tab:main}, the final efficient block reports accuracies stated in
  each method's own paper under different backbones, frame counts, and protocols,
  so these rows are partial and receive no aggregate.}
\label{tab:mainfull}
\centering
\setlength{\tabcolsep}{3pt}
\renewcommand{\arraystretch}{0.95}
\resizebox{\textwidth}{!}{%
\begin{tabular}{@{}lcccccccccccc@{}}
\toprule
\textbf{Model}
 & \benchcell{NextQA}{test \citep{xiao2021nextqa}}
 & \benchcell{PerceptionTest}{test \citep{patraucean2023perceptiontest}}
 & \benchcell{MVBench}{test \citep{li2024mvbench}}
 & \benchcell{Tomato}{test \citep{shangguan2025tomato}}
 & \benchcell{Short QA avg.}{}
 & \benchcell{Video-MME}{test \citep{fu2025videomme}}
 & \benchcell{Video-MME-Sub}{test \citep{fu2025videomme}}
 & \benchcell{LongVideoBench}{val \citep{wu2024longvideobench}}
 & \benchcell{MLVU}{test MCQ \citep{zhou2025mlvu}}
 & \benchcell{LVBench}{test \citep{wang2025lvbench}}
 & \benchcell{Long QA avg.}{}
 & \benchcell{Average}{}\\
\midrule
\multicolumn{13}{@{}l}{\emph{Proprietary models, API only}}\\
GPT-5 \citep{openai2025gpt5}
 & 86.3 & 79.4 & 74.1 & 53.0 & 73.2
 & 83.3 & 86.9 & 72.6 & 77.7 & 65.2 & 77.1
 & 75.4\\
GPT-5 mini \citep{openai2025gpt5}
 & 83.2 & 72.0 & 66.5 & 44.1 & 66.5
 & 77.3 & 82.3 & 69.7 & 69.1 & 54.7 & 70.6
 & 68.8\\
Gemini 3 Pro \citep{google2025gemini3}
 & 84.3 & 77.6 & 70.4 & 48.3 & 70.2
 & 88.6 & 87.5 & 75.9 & 75.7 & 77.0 & 80.9
 & 76.1\\
Gemini 2.5 Pro \citep{comanici2025gemini25}
 & 85.3 & 78.4 & 70.6 & 48.6 & 70.7
 & 87.8 & 87.8 & 76.8 & 81.5 & 75.7 & 81.9
 & 76.9\\
Gemini 2.5 Flash \citep{comanici2025gemini25}
 & 81.8 & 74.7 & 67.0 & 39.1 & 65.7
 & 84.2 & 84.2 & 73.1 & 75.1 & 64.9 & 76.3
 & 71.6\\
Claude Sonnet 4.5 \citep{anthropic2025sonnet45}
 & 79.2 & 64.3 & 62.1 & 39.6 & 61.3
 & 74.2 & 80.5 & 65.1 & 64.0 & 50.5 & 66.9
 & 64.4\\
\midrule
\multicolumn{13}{@{}l}{\emph{Open models}}\\
InternVL3.5-4B \citep{wang2025internvl35}
 & 80.3 & 68.1 & 71.2 & 26.8 & 61.6
 & 65.4 & 68.6 & 60.8 & 52.0 & 43.2 & 58.0
 & 59.6\\
InternVL3.5-8B \citep{wang2025internvl35}
 & 81.7 & 72.7 & 72.1 & 24.6 & 62.8
 & 66.0 & 68.6 & 62.1 & 53.2 & 43.4 & 58.7
 & 60.5\\
Qwen3-VL-4B \citep{bai2025qwen3vl}
 & 81.4 & 70.7 & 68.9 & 31.8 & 63.2
 & 69.3 & 74.0 & 62.8 & 58.4 & 56.2 & 64.1
 & 63.7\\
Qwen3-VL-8B \citep{bai2025qwen3vl}
 & 83.4 & 72.7 & 68.7 & 35.7 & 65.1
 & 71.4 & 75.2 & 62.4 & 57.6 & 58.0 & 64.9
 & 65.0\\
Keye-VL-1.5-8B \citep{yang2025keyevl}
 & 75.8 & 64.2 & 56.9 & 33.0 & 57.5
 & 73.0 & 76.2 & 66.0 & 53.8 & 42.8 & 62.4
 & 60.2\\
GLM-4.1V-9B \citep{hong2025glm41v}
 & 81.3 & 74.2 & 68.4 & 30.0 & 63.5
 & 68.2 & 75.6 & 65.7 & 56.6 & 44.0 & 62.0
 & 62.7\\
MiniCPM-V-4.5-8B \citep{yu2025minicpmv45}
 & 78.8 & 70.9 & 60.5 & 29.8 & 60.0
 & 67.9 & 73.5 & 63.9 & 60.6 & 50.4 & 63.3
 & 61.8\\
Eagle2.5-8B \citep{chen2025eagle25}
 & 85.0 & 81.0 & 74.8 & 31.0 & 68.0
 & 72.4 & 75.7 & 66.4 & 60.4 & 50.9 & 65.2
 & 66.4\\
PLM-3B \citep{cho2025perceptionlm}
 & 83.4 & 79.3 & 74.7 & 30.9 & 67.1
 & 54.9 & 59.4 & 57.9 & 48.4 & 40.4 & 52.2
 & 58.8\\
PLM-8B \citep{cho2025perceptionlm}
 & 84.1 & 82.7 & 77.1 & 33.2 & 69.3
 & 58.3 & 65.4 & 56.9 & 52.6 & 44.5 & 55.5
 & 61.6\\
LLaVA-Video-7B \citep{zhang2024llavavideo}
 & 83.2 & 68.8 & 58.6 & 24.9 & 58.9
 & 63.3 & 69.7 & 58.2 & 52.8 & 44.2 & 57.6
 & 58.2\\
LLaVA-OneVision-7B \citep{li2024llavaonevision}
 & 79.4 & 57.1 & 56.7 & -- & --
 & 58.2 & 61.5 & 56.4 & -- & -- & --
 & --\\
VideoChat-Flash-7B \citep{li2024videochatflash}
 & 85.5 & 76.5 & 74.0 & 32.5 & 67.1
 & 65.3 & 69.7 & 64.7 & 56.0 & 48.2 & 60.8
 & 63.6\\
Molmo2-4B \citep{molmo2team2026}
 & 85.5 & 81.3 & 75.1 & 39.8 & 70.4
 & 69.6 & 75.7 & 68.0 & 63.0 & 53.9 & 66.0
 & 68.0\\
Molmo2-8B \citep{molmo2team2026}
 & 86.2 & 82.1 & 75.9 & 39.6 & 71.0
 & 69.9 & 75.8 & 67.5 & 60.2 & 52.8 & 65.2
 & 67.8\\
Molmo2-O-7B \citep{molmo2team2026}
 & 84.3 & 79.6 & 74.8 & 36.2 & 68.7
 & 64.9 & 69.2 & 63.7 & 55.2 & 49.6 & 60.5
 & 64.2\\
\midrule
\multicolumn{13}{@{}l}{\emph{Efficient video language models} (re-evaluated at a common 256-frame budget)}\\
BIMBA \citep{islam2025bimba}
 & 80.9 & 65.8 & 59.6 & 24.0 & 57.6
 & 63.3 & 66.4 & 57.7 & \textbf{53.9} & 43.3 & 56.9
 & 57.2\\
Vamba \citep{ren2025vamba}
 & 77.5 & \textbf{69.4} & 59.6 & 26.2 & 58.2
 & 58.1 & 61.2 & 52.8 & 45.7 & 41.2 & 51.8
 & 54.6\\
VideoNSA \citep{song2025videonsa}
 & 77.1 & 56.7 & 57.0 & 22.7 & 53.4
 & 59.4 & 61.9 & 52.2 & 41.3 & 41.8 & 51.3
 & 52.2\\

\method{} ($k = 16$)
 & \textbf{81.8} & 66.7 & \textbf{62.8} & \textbf{30.9} & \textbf{60.6}
 & \textbf{66.2} & \textbf{67.5} & \textbf{60.4} & 52.0 & \textbf{44.0} & \textbf{58.0}
 & \textbf{59.1}\\
\method{} ($k = 12$)
 & 79.6 & 66.2 & 60.9 & 29.3 & 59.0
 & 64.1 & 66.5 & 59.6 & 52.5 & 42.0 & 56.9
 & 57.9\\
\method{} ($k = 8$)
 & 80.0 & 65.4 & 60.6 & 28.7 & 58.7
 & 63.8 & 66.3 & 58.8 & 51.9 & 40.0 & 56.2
 & 57.3\\
\method{} ($k = 6$)
 & 79.9 & 63.4 & 60.0 & 28.4 & 57.9
 & 62.0 & 65.5 & 58.6 & 50.1 & 41.3 & 55.5
 & 56.6\\
\addlinespace[2pt]
\multicolumn{13}{@{}l}{\emph{Additional video language models, scores as reported by their original papers} (mixed backbones and protocols, partial coverage)}\\
LLaVA-Mini \citep{zhang2025llavamini}
 & -- & -- & 44.5 & -- & --
 & -- & -- & -- & -- & -- & --
 & --\\

\bottomrule
\end{tabular}%
}
\end{table*}

\begin{table}[t]
\caption{Latency of the uncompressed Qwen3-VL-8B backbone and \method{} on one
B200 GPU in bfloat16. All timings are batch size one: per example updates cannot be
batched across examples. For \method{}, prefill and time to first token include the
compressor, the hypernetwork forward, and the cost of merging the generated update.
Prefill excludes the vision encoder, while time to first token includes it. Decoding
is measured after merging the generated update.}
\label{tab:latency-grid}
\centering
\small
\setlength{\tabcolsep}{4pt}
\begin{tabular*}{\textwidth}{@{\extracolsep{\fill}}rrrrrrrrr@{}}
\toprule
& & \multicolumn{2}{c}{prefill (ms)} & \multicolumn{2}{c}{TTFT (ms)} & & \multicolumn{2}{c}{decode (ms/token)}\\
\cmidrule(lr){3-4}\cmidrule(lr){5-6}\cmidrule(l){8-9}
$k$ & frames & base & \method{} & base & \method{} & encoder & base & \method{}\\
\midrule
16 & 128  & 180 & 54  & 274  & 162  & 94  & 18.8 & 19.7\\
16 & 256  & 369 & 110 & 550  & 323  & 181 & 18.9 & 18.8\\
16 & 512  & 851 & 228 & 1216 & 659  & 365 & 21.1 & 18.9\\
16 & 1024 & 2086 & 504 & 2819 & 1367 & 733 & 21.1 & 18.2\\
\midrule
6  & 128  & 183 & 30  & 276  & 136  & 93  & 19.0 & 19.1\\
6  & 256  & 359 & 57  & 550  & 273  & 192 & 18.5 & 19.2\\
6  & 512  & 824 & 111 & 1199 & 542  & 375 & 20.6 & 21.9\\
6  & 1024 & 2152 & 238 & 2900 & 1099 & 748 & 20.0 & 19.1\\
\bottomrule
\end{tabular*}
\end{table}

\section{What the generated update encodes}
\label{sec:vniah}

The ablations of Section~\ref{sec:results-weight} bound the contribution of the generated
low rank update on benchmark questions, where a large part of each answer is recoverable
from the question text alone. To measure the update's information content without that
component, we adapt the needle in a haystack protocol of \citet{cheng2025vilamp} so that
the answer is a target the question text cannot supply.

\paragraph{Protocol.}
A needle word (drawn from a 32 word vocabulary) is rendered across 32 consecutive frames
and spliced into a LongVideoBench haystack at one of five depths (0 to 100\%), then probed
as a four way multiple choice. The word is determined by neither haystack nor question, so
chance is $25\%$. Haystacks run 256 to 2048 frames (the needle falls from $12.5\%$ to
$1.6\%$ of the input), 120 items per model and length. Options are rotated to a uniform key
and the per frame resolution is fixed at 79{,}872\,px, so letter bias and shrinking frames
do not confound the measurement.

\begin{figure}[t]
\centering
\includegraphics[width=\textwidth]{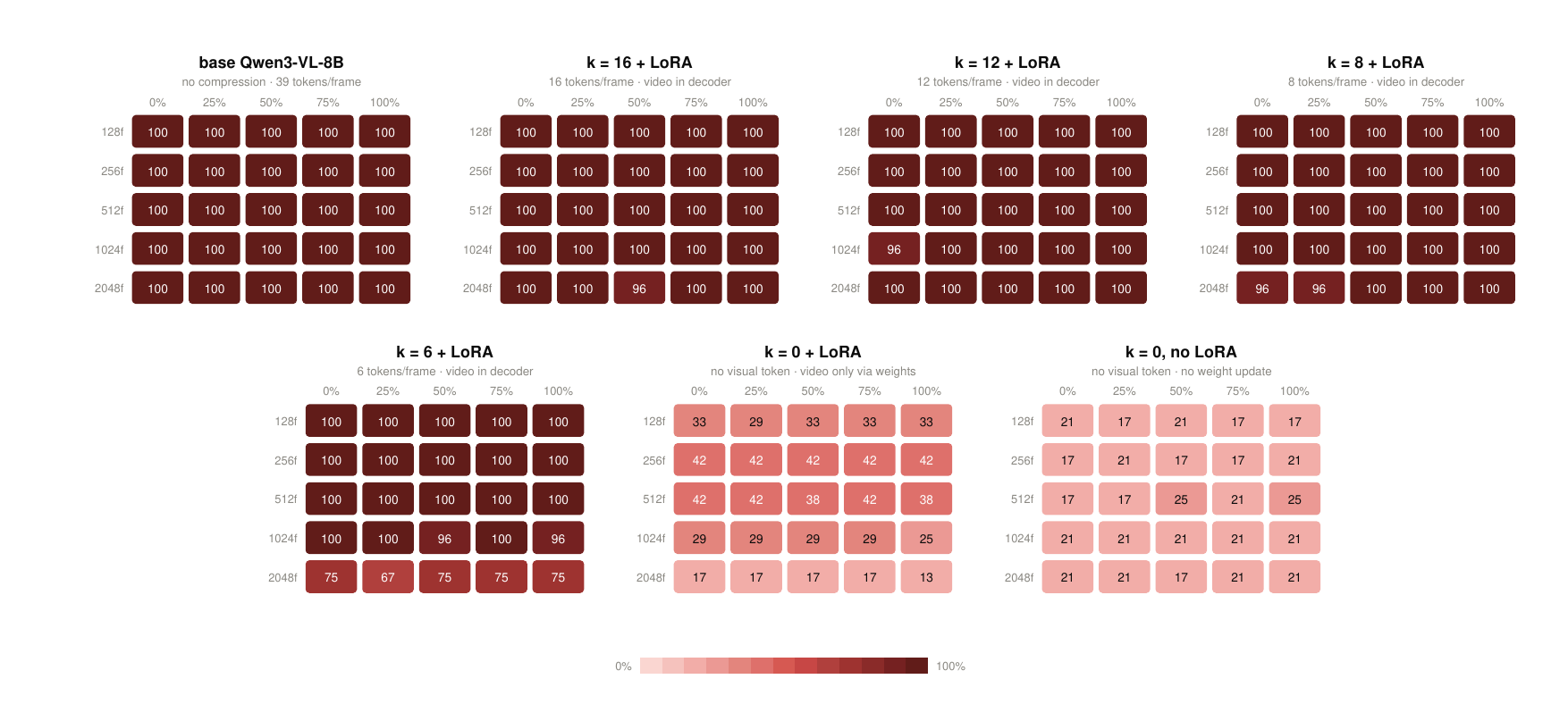}
\caption{\textbf{Video needle in a haystack under decoder input ablation.} Retrieval
accuracy (\%) for a planted word card needle across five insertion depths (columns of each
grid) and three haystack lengths (rows). Darker cells are higher. The uncompressed backbone
and \method{} at carrier budgets $k\in\{16,12,8,6\}$ keep the video in the decoder input.
The two $k{=}0$ grids remove every visual token, so the generated update is the only path
from video to answer, and differ only in whether that update is applied. Chance is
$25\%$.}
\label{fig:vniah}
\end{figure}

\paragraph{Contraction preserves the needle, the generated update does not carry it.}
Figure~\ref{fig:vniah} separates the two paths by which the video can reach the answer. With
the contracted video present in the decoder input sequence, every width from $k{=}16$ down
to $k{=}8$ matches the uncompressed model at every position and every sequence length, at
$8/39$ of the visual tokens. The $k{=}6$ width is the first to lose the needle. It holds
through 1024 frames and falls to $73.3$ at 2048. Accuracy is flat across the five positions
for every model, so the loss is a function of the fraction of the input the needle occupies
and not of where it sits.

The $k{=}0$ checkpoint, trained with no visual token in the decoder input sequence, isolates
the other path: the generated update is the only route from the video to the output, and its
two rows differ solely by whether that update is applied. At 256 frames the checkpoint
scores $41.7$ with the update against $18.3$ without it, a paired difference of $+23.3$
points (38/10 discordant, exact McNemar $p = 6.2 \times 10^{-5}$), and $+15.8$ points over
an update generated from a different video ($p = 0.003$). The effect does not survive
dilution: $+7.5$ points at 1024 frames ($p = 0.14$) and $-4.2$ at 2048 ($p = 0.40$). Run on
$k{=}6$ through $k{=}16$, the same contrast is null at every length.

\end{document}